\documentclass[11pt]{article}
\usepackage[preprint]{acl}

\usepackage[T1]{fontenc}
\usepackage{microtype}
\usepackage{inconsolata}
\usepackage{times}  
\usepackage{amsmath}
\usepackage{amsfonts}
\usepackage{amssymb} 
\usepackage{mathtools}
\usepackage{bm}
\usepackage{bbm}
\usepackage{nicefrac}

\usepackage{graphicx}
\usepackage{wrapfig}
\usepackage{placeins}
\usepackage{float}
\usepackage{adjustbox}
\usepackage{booktabs}
\usepackage{tabularx}
\usepackage{array}
\usepackage{multirow}
\usepackage{colortbl}
\newcolumntype{Y}{>{\raggedright\arraybackslash}X}

\usepackage{algorithm}
\usepackage{algpseudocode}

\usepackage{enumitem}
\setlist[itemize]{leftmargin=*, itemsep=2pt, topsep=2pt}
\setlist[enumerate]{leftmargin=*, itemsep=2pt, topsep=2pt}

\usepackage{xcolor}
\usepackage[most]{tcolorbox}
\definecolor{highlightpurple}{HTML}{E6E6FA}
\definecolor{MyCustomBlue}{HTML}{92CE6A}
\definecolor{MyCustomPink}{HTML}{C7513E}
\definecolor{SoftGray}{HTML}{F5F5F5}

\usepackage{url}
\usepackage{tikz}
\newcommand{\circnum}[1]{%
  \tikz[baseline=(char.base)]{
    \node[
      shape=circle,
      fill=black,
      text=white,
      inner sep=0.9pt,
      font=\scriptsize\sffamily\bfseries
    ] (char) {#1};
  }%
}
\title{AlphaToken: Decoupling Adaptation and Stability for Path-Aware Response Token Valuation in LLM Post-Training}

\author{
  \textbf{Qing Liu},
  \textbf{Ou Wu\thanks{Corresponding author.}},
  \textbf{Yi Du}
\\
\\
 Hangzhou Institute for Advanced Study, \\ University of Chinese Academy of Sciences\\
 \vspace{0.1cm}
\texttt{liuqing25@mails.ucas.ac.cn}
  \quad
\texttt{wuou@ucas.ac.cn}
}

\begin{document}
\maketitle

\begin{abstract}
Token selection is pivotal for effective LLM post-training.
However, existing methods mostly rely on local heuristics and rarely formulate token selection as a principled valuation of individual response tokens.
We introduce \textbf{AlphaToken}, a response token valuation framework that decouples valuation into \textbf{adaptation} (promoting target-task learning) and \textbf{stability} (preserving pre-trained capabilities), and makes each objective \textbf{path-aware} by combining the direct-path signal from local token gradients with the downstream causal-path signal in autoregressive generation.
Since retention data are typically unavailable, AlphaToken approximates stability via a \textbf{Fisher-drift proxy} anchored at the pre-trained reference model.
For efficient computation, we extend Ghost Dot-Product to token-level valuation.
AlphaToken masks low-value response tokens during fine-tuning and preference optimization, concentrating training signals on more valuable positions. Experiments show that AlphaToken improves post-training performance and mitigates catastrophic forgetting. Code at \url{https://anonymous.4open.science/r/AlphaToken-58A}.
\end{abstract}

\section{Introduction}

Large language models~(LLMs) are increasingly adapted through post-training to meet evolving domain needs and alignment goals. In practice, supervised fine-tuning (SFT) and preference optimization (PO) have become the dominant adaptation paradigms~\citep{ouyang2022training,rafailov2023direct}. Recent work has shown that data utility matters for adaptation, from sample valuation and selection \citep{liu2024selectit,wang2025data} to emerging token-wise methods that filter tokens for efficiency or robustness \citep{pang2025token,simoulin2024memory}.

Classical formulations build on influence functions and Data Shapley, which estimate each training sample's marginal contribution to validation performance~\citep{koh2017understanding,ghorbani2019data}. Recent work has improved the scalability of this line for LLM training pipelines. In-Run Data Shapley estimates Shapley-style contributions within a single training trajectory~\citep{wang2025data}, trajectory-aware influence estimation accounts for when a datum appears during optimization~\citep{wang2025capturing}, and GREATS performs online selection of high-quality batches during LLM training~\citep{wang2024greats}. 
While these advances substantially improve scalability, they remain sample-level and cannot capture the heterogeneous contributions that individual response tokens make within a single sequence.

Emerging token selection methods begin to address this issue~\citep{pang2025token,simoulin2024memory}. Existing methods typically rely on teacher--student loss gaps, low-perplexity masking, or alignment-specific token weighting~\citep{pang2025token,wu2025mitigating,yang2026tokenimportance}. These criteria are efficient but largely heuristic and single-objective. Although Shapley-style token attribution methods provide a principled view of token importance, they mainly target input-side context attribution and model interpretation~\citep{horovicz2024tokenshap,xiao2025tokenshapley}. This leaves open the need for a principled valuation framework that measures how each response token contributes to LLM post-training.

We propose \textbf{AlphaToken}, a fine-grained response token valuation framework for LLM post-training. It decouples the valuation objective into target adaptation and retention stability. AlphaToken achieves path-aware valuation by jointly modeling a response token's direct-path update signal and its downstream causal-path influence on later predictions in autoregressive generation. Since retention data are typically unavailable in practical post-training, it estimates retention stability with a Fisher-drift proxy anchored at the pre-trained reference model, enabling stability-aware valuation when no retention examples are available. For efficient computation, we extend Ghost Dot-Product~(GDP) to token-level valuation. The composite token values are used to mask low-value response tokens during fine-tuning and preference optimization, concentrating training on positions that better support adaptive and stable post-training. Our key contributions are summarized as follows:

\begin{itemize}[leftmargin=*]
\item We formulate response token valuation by decoupling adaptation and retention stability, and make it path-aware by modeling local update and downstream causal signals. We replace inaccessible retention gradients with a Fisher-weighted drift proxy anchored at the pre-trained model.

\item Building on GDP for sample-level gradient alignment, we extend it to token-level valuation by supporting causal cross-position alignment through a Value-Propagation approximation and Fisher-drift stability through a new Activation-Parameter contraction.
\item Using composite values, we develop a token selection method for post-training scenarios that masks low-utility response tokens in SFT and PO, improving Overall by at least $1.54$ points.
\end{itemize}

\section{Related Work}
\subsection{Data Valuation and Influence Estimation.}
Data valuation studies each training example's contribution to a target metric. Classical approaches center on Data Shapley and scalable variants, including Monte Carlo estimation, acceleration, and semivalue relaxations such as Beta Shapley and Banzhaf values \citep{ghorbani2019data,jia2019towards,mitchell2022sampling,okhrati2021multilinear,kwon2022beta,wang2023data,li2023robust}. Parallel lines pursue cheaper influence approximations, Bayesian uncertainty formulations, and scalable surrogates such as KNN, distributional, and federated variants \citep{koh2017understanding,basu2021influence,nguyen2023bayesian,jia2019efficient,ghorbani2020distributional,wang2020principled}. Recent work models trajectory dependence through data value embeddings and adaptive reference points for real-time valuation \citep{wang2025capturing,xu2025liveval}. Although some studies extend Shapley attribution to token-level prompt interpretation, applying such ideas directly to training-time token valuation remains computationally challenging \citep{horovicz2024tokenshap,xiao2025tokenshapley}. In-Run Data Shapley mitigates retraining cost by accumulating per-step gradient interactions along a single trajectory \citep{wang2025data}. In contrast, we estimate objective-aligned token values in one run through gradient alignment with validation utility and a data-free stability proxy.

\subsection{Token Selection \& Reweighting.}
Token-level data selection and reweighting offer an alternative to sample-level filtering for improving post-training efficiency and robustness. Most methods rely on local token heuristics, such as loss- or probability-based scoring, excess-loss and attention-based relevance, regime-aware joint pruning, and perplexity-based masking; similar ideas appear in pretraining and multimodal training \citep{pang2025token,liu2026profit,qin2026sstoken,nagaraj2025trim,wang2025winning,wu2025mitigating,lin2024not,peng2025large}. Other lines explore memory-efficient selective backprop, task-structure-aware token weighting, and alignment-oriented token reweighting \citep{simoulin2024memory,shi2025rethinking,yang2026tokenimportance,le2025token}. In contrast, our method scores tokens using gradient alignment to validation utility and a data-free stability proxy, producing objective-consistent signals for selection.

\begin{figure*}[t]
\centering
\includegraphics[width=\textwidth]{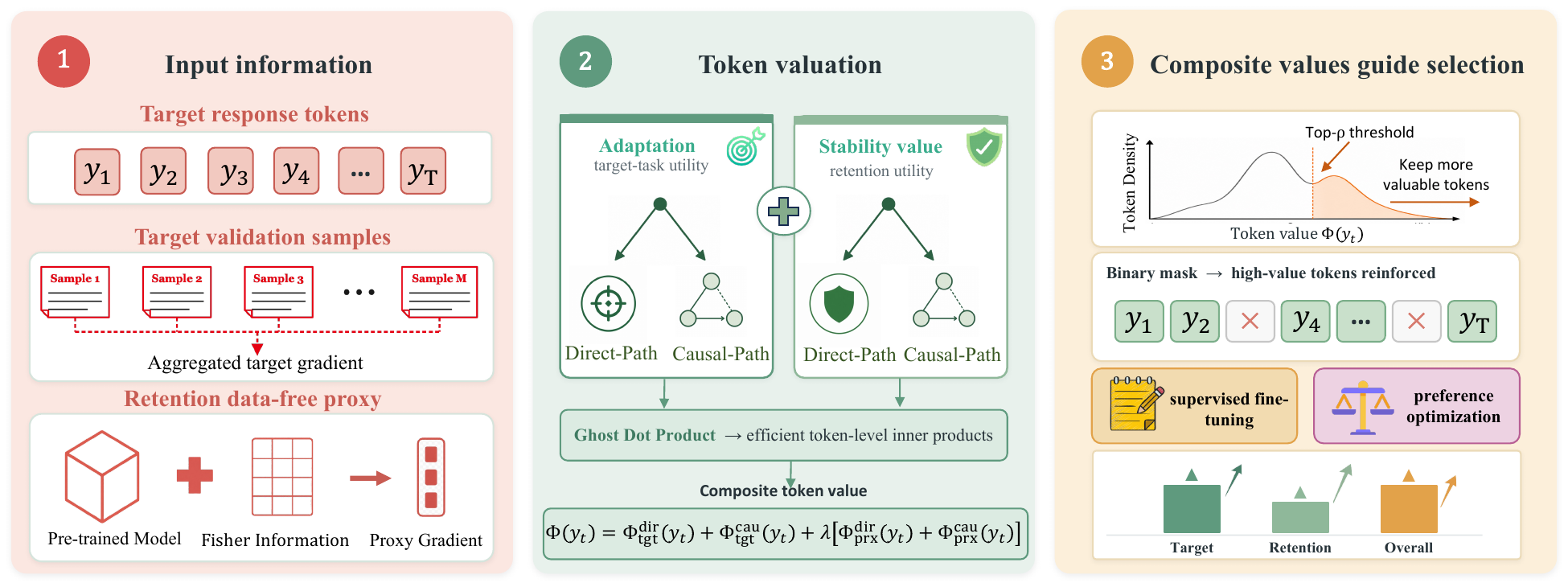}
\vspace{-8mm}
\caption{Overview of the AlphaToken framework for selecting high-value response tokens in post-training.}
\label{fig:framework}
\vspace{-4mm}
\end{figure*}

\section{Methodology}
\label{sec:Methodology}

\subsection{Preliminaries and Problem Formulation}
Consider a causal language model parameterized by $\bm{\theta}$. Let a training instance consist of a prompt $\bm{x}=[x_1,\dots,x_P]$ and a response $\bm{y}=[y_1,\dots,y_T]$. The training loss $\mathcal{L}_{\text{train}}$ can be either the SFT cross-entropy loss or a preference-optimization loss, with DPO as a representative instantiation. In both instantiations, the loss can be expressed through response-token contributions, allowing us to define a token-level loss $\ell_t$ for SFT and a token-level pseudo-loss $\ell_{t}^{\pm}$ for DPO. We assign a scalar valuation $\Phi(y_t)$ to each response token to quantify its marginal generalization contribution.
Let $\mathcal{D}_{\text{tgt}}^{\text{val}}$ and $\mathcal{D}_{\text{ret}}^{\text{val}}$ denote target-domain and retention validation sets. Balancing target adaptation and retention stability, we define a composite objective
\begin{equation}
\resizebox{0.88\linewidth}{!}{$
\displaystyle
J_{\text{val}}(\bm{\theta}) = J_{\text{tgt}}(\bm{\theta}; \mathcal{D}_{\text{tgt}}^{\text{val}}) + \lambda \cdot J_{\text{ret}}(\bm{\theta}; \mathcal{D}_{\text{ret}}^{\text{val}}),
$}
\label{eq:composite_val}
\end{equation}
where $J_{\text{tgt}}$ and $J_{\text{ret}}$ denote target and retention validation losses, and $\lambda$ controls retention strength. Building on the sample-level in-run Shapley view of \citet{wang2025data}, we extend the argument to token-level masking in App.~\ref{app:optimal_masking_to_shapley} and define each response token's value as proportional to the alignment between its training gradient and the composite validation gradient. The valuation thus decomposes into plasticity and stability terms:
\begin{equation}
\resizebox{0.88\linewidth}{!}{$
\displaystyle
\Phi(y_t) \propto
\big\langle \nabla_{\bm{\theta}} \ell_t,\;
\nabla_{\bm{\theta}} J_{\text{tgt}} \big\rangle
+ \lambda
\big\langle \nabla_{\bm{\theta}} \ell_t,\;
\nabla_{\bm{\theta}} J_{\text{ret}} \big\rangle.
$}
\label{eq:grad_alignment}
\end{equation}

Directly applying Eq.~\eqref{eq:grad_alignment} to token valuation faces
three obstacles.
\textbf{First}, the local gradient $\nabla_{\bm{\theta}}\ell_t$ undercounts
$y_t$'s influence, omitting its effect on later predictions through
autoregressive context.
\textbf{Second}, $\nabla_{\bm{\theta}}J_{\mathrm{ret}}$ depends on the
retention validation set $\mathcal{D}_{\mathrm{ret}}^{\mathrm{val}}$, which
is typically unavailable in post-training.
\textbf{Third}, the parameter-space gradient inner products must be computed
per response token, prohibitive if token-level gradients are formed.

\subsection{Causal Gradient Decomposition}
\label{sec:causal_decomp}
In autoregressive Transformers, a token $y_t$ plays two roles. As a label, it generates the immediate loss $\ell_t$ at position $t$. As context, its hidden state influences all future tokens $y_k$ ($k>t$) through self-attention. Routing the loss of each future token backward through the layer-$l$ cross-position transfer from position $t$ to position $k$ and isolating the self-term $k{=}t$ yields
\begin{equation}
\resizebox{0.88\linewidth}{!}{$
\displaystyle
\nabla_{\bm{\theta}} \ell_t^{\text{tot}} = \nabla_{\bm{\theta}} \ell_t + \!\sum_{k=t+1}^{T}\!\sum_{l=1}^{L}\! \frac{\partial \ell_k}{\partial \bm{h}_k^{l}} \frac{\partial \bm{h}_k^{l}}{\partial \bm{h}_t^{l-1}} \frac{\partial \bm{h}_t^{l-1}}{\partial \bm{\theta}}.
$}
\label{eq:causal_decomp}
\end{equation}

Eq.~\eqref{eq:causal_decomp} isolates the dominant Value-Propagation channel
from $t$ to $k$, derived in App.~\ref{app:causal_decomp} by routing each future
loss through the cross-position Jacobian and decomposing its attention derivative
into the value path. The same
chain-rule argument applies to both SFT and DPO losses.

\subsection{Data-Free Retention Gradient Proxy}
The composite objective in Eq.~\eqref{eq:grad_alignment} requires the retention validation gradient $\nabla_{\bm{\theta}} J_{\text{ret}}$. In realistic post-training, however, the pre-training corpus is unavailable to downstream practitioners due to licensing constraints and closed-source release policies~\cite{kirkpatrick2017overcoming,sanyal2025upweighting}. 
Thus, we adopt a setting where retention data are unavailable, with access limited to the pre-trained parameters $\bm{\theta}_{\text{ref}}$ and the target dataset $\mathcal{D}_{\text{tgt}}$.

Let $\mathcal{L}_{\text{ret}}(\bm{\theta})$ denote the unobserved retention loss. Expanding around $\bm{\theta}_{\text{ref}}$:
\vspace{-0.05in}
\begin{equation}
\resizebox{0.88\linewidth}{!}{$
\displaystyle
\mathcal{L}_{\text{ret}}(\bm{\theta}) \approx
\mathcal{L}_{\text{ret}}(\bm{\theta}_{\text{ref}})
+ \tfrac{1}{2}(\bm{\theta}-\bm{\theta}_{\text{ref}})^{\!\top} \bm{H}_{\text{ret}} (\bm{\theta}-\bm{\theta}_{\text{ref}}),
$}
\label{eq:retention_taylor}
\vspace{-0.05in}
\end{equation}
where the first-order term is small to the extent that $\bm{\theta}_{\text{ref}}$ approximates a stationary point of $\mathcal{L}_{\text{ret}}$ (we bound the residual in App.~\ref{app:retention_proxy_bound}), and $\bm{H}_{\text{ret}}$ coincides with the Fisher information at $\bm{\theta}_{\text{ref}}$ under the maximum-likelihood calibration condition~\cite{martens2020new}. Replacing $\bm{H}_{\text{ret}}$ with a model-side Fisher $\bm{F}_{\text{ref}}$ that requires no retention data yields the data-free retention proxy:
\vspace{-0.05in}
\begin{equation}
J_{\text{prx}}(\bm{\theta}) =
\tfrac{1}{2}(\bm{\theta}-\bm{\theta}_{\text{ref}})^{\!\top}
\bm{F}_{\text{ref}}\,
(\bm{\theta}-\bm{\theta}_{\text{ref}}).
\label{eq:fisher_proxy}
\vspace{-0.05in}
\end{equation}

The Fisher information characterizes the likelihood geometry of $\bm{\theta}_{\text{ref}}$, depending on the reference model and prompt distribution rather than retention labels/samples. 
Here, $\mathcal{X}$ denotes the prompt distribution for Monte-Carlo Fisher construction, instantiated by prompts from $\mathcal{D}_{\text{tgt}}$.
To avoid the bias of the empirical Fisher far from a local optimum~\cite{kunstner2019limitations}, we adopt the Monte-Carlo Fisher with labels self-sampled from $p_{\bm{\theta}_{\text{ref}}}$:
\begin{equation*}
\resizebox{0.99\linewidth}{!}{$
\displaystyle
\bm{F}_{\text{ref}} = \mathrm{Diag}\!\left(
\mathbb{E}_{\bm{x}\sim\mathcal{X},\,\tilde{\bm{y}}\sim p_{\bm{\theta}_{\text{ref}}}(\cdot|\bm{x})}
\big[\, \nabla_{\bm{\theta}}\log p_{\bm{\theta}_{\text{ref}}}(\tilde{\bm{y}}|\bm{x})^{\odot 2} \,\big]
\right),
$}
\label{eq:mc_fisher}
\end{equation*}
$\bm{F}_{\text{ref}}$ is computed once before training and cached.

Differentiating Eq.~\eqref{eq:fisher_proxy} gives
\begin{equation}
\resizebox{0.8\linewidth}{!}{$
\bm{g}_{\text{prx}}\;\triangleq\; \nabla_{\bm{\theta}} J_{\text{prx}}(\bm{\theta})\;=\;
\bm{F}_{\text{ref}}(\bm{\theta} - \bm{\theta}_{\text{ref}}),
$}
\label{eq:gproxy}
\end{equation}
which acts as a virtual retention validation gradient: substituting $\bm{g}_{\text{prx}}$ for $\nabla_{\bm{\theta}} J_{\text{ret}}$ in Eq.~\eqref{eq:grad_alignment} preserves the alignment formulation while removing retention data dependency. Tokens whose training gradients align with $\bm{g}_{\text{prx}}$ contract the Fisher-weighted drift from $\bm{\theta}_{\text{ref}}$ and are therefore retention-friendly. The corresponding retention valuation reads
\begin{equation}
\Phi_{\text{prx}}(y_t)\;\triangleq\;
\big\langle \nabla_{\bm{\theta}}\ell_t^{\text{tot}},\, \bm{g}_{\text{prx}}\big\rangle.
\label{eq:phi_proxy}
\end{equation}
By the linearity of the inner product and the direct/causal split of $\nabla_{\bm{\theta}}\ell_t^{\text{tot}}$ in Eq.~\eqref{eq:causal_decomp}, $\Phi_{\text{prx}}(y_t)=\Phi_{\text{prx}}^{\text{dir}}(y_t)+\Phi_{\text{prx}}^{\text{cau}}(y_t)$.


\subsection{Ghost Dot-Product Approximation}
\label{sec:efficient}
Realizing Eq.~\eqref{eq:grad_alignment} requires token-level inner products for direct target alignment, causal future-token alignment, and retention alignment with $\bm{g}_{\text{prx}}$. 
Naively materializing per-token parameter gradients costs $\mathcal{O}(T|\bm{\theta}|)$, which is prohibitive at LLM scale. 
We compute all three with a unified GDP family: the direct activation--activation (A--A) form adapts the example-level construction of~\citet{wang2025data} to tokens, while the causal A--A form and the activation--parameter (A--P) proxy form are new. 
Scores are aggregated over the last \(K\) layers \(\mathcal{S}\).

\textbf{Direct token alignment (A--A).}
For a linear layer $\bm{W}^{l}$, the token-level loss yields a rank-1 weight gradient $\nabla_{\bm{W}^{l}}\ell_t = \bm{\delta}_t^{l}\!\otimes\!\bm{h}_t^{l}$ with input activation $\bm{h}_t^{l}$ and output error $\bm{\delta}_t^{l}\!\triangleq\!\partial\ell_t/\partial\bm{h}_t^{l}$. Following~\citet{wang2025data}, the Frobenius inner product factorizes into two activation-space dot products. Concatenating a small fixed validation batch with the training batch in a single backward pass~\citep{wang2024greats} yields per-token training signals $(\bm{\delta}_t^{l},\bm{h}_t^{l})$ and per-validation-token target signals $(\bm{\delta}_{v}^{l},\bm{h}_{v}^{l})_{v\in\mathcal{V}_{\text{tgt}}}$ from the same cache, where $\mathcal{V}_{\text{tgt}}$ denotes the held-out target validation token set. Summing over layers and averaging over $\mathcal{V}_{\text{tgt}}$, the direct target valuation reads
\begin{equation}
\vspace{-0.02in}
\resizebox{0.86\linewidth}{!}{$
\displaystyle
\Phi_{\text{tgt}}^{\text{dir}}(y_t)
=
\frac{1}{|\mathcal{V}_{\text{tgt}}|}\sum\nolimits_{v\in\mathcal{V}_{\text{tgt}}}\sum\nolimits_{l\in\mathcal{S}} \big\langle \bm{\delta}_t^{l},\,\bm{\delta}_{v}^{l}\big\rangle\, \big\langle \bm{h}_t^{l},\,\bm{h}_{v}^{l}\big\rangle.
$}
\label{eq:phi_tgt_dir}
\end{equation}

\textbf{Causal token alignment (A--A causal).}
The cross-position Jacobian is dense; we approximate it by retaining only linear Value-Propagation through $\bm{W}_V^{l}$ and omitting nonlinear Score-Propagation, whose magnitude decays as $\mathcal{O}(1/\sqrt{d_h})$ under saturated attention, where $d_h$ is the attention-head dimension (App.~\ref{app:value_path_bound}). This yields
$\nabla_{\bm{W}_V^{l}}\,\ell_{t\to k}\!\approx\! \alpha_{t\to k}^{l}(\bm{\delta}_k^{l}\!\otimes\!\bm{h}_t^{l-1})$ with attention weight $\alpha_{t\to k}^{l}$. The activation-space factorization, averaged over $\mathcal{V}_{\text{tgt}}$, gives the causal target valuation
\begin{equation}
\resizebox{0.88\linewidth}{!}{$
\displaystyle
\Phi_{\text{tgt}}^{\text{cau}}(y_t)
=
\frac{1}{|\mathcal{V}_{\text{tgt}}|}\sum_{v\in\mathcal{V}_{\text{tgt}}}\sum_{l\in\mathcal{S},\,k} \alpha_{t\to k}^{l} \big\langle \bm{\delta}_k^{l}, \bm{\delta}_{v}^{l}\big\rangle \big\langle \bm{h}_t^{l-1},\bm{h}_{v}^{l-1}\big\rangle,
$}
\label{eq:phi_tgt_causal}
\end{equation}
with $k\!\in\!(t,\min(t\!+\!W,T)]$ for causal window $W$.

\textbf{Retention proxy (A--P).}
The retention gradient $\bm{g}_{\text{prx}}$ is a fixed parameter-space vector and is not the back-propagation output of any sample-level loss; it admits no activation-only factorization. We instead contract the rank-1 token gradient against $\bm{g}_{\text{prx}}$ directly. For each linear layer, define
\begin{equation}
\bm{V}^{l}\;\triangleq\;\bm{F}_{\bm{W}^{l}}\,\odot\,\big(\bm{W}^{l} - \bm{W}_{\text{ref}}^{l}\big),
\label{eq:V_layer}
\end{equation}
the layer-$l$ slice of $\bm{g}_{\text{prx}}$, where $\bm{F}_{\bm{W}^{l}}\!\triangleq\!\mathrm{Diag}(\bm{F}_{\text{ref}})|_{\bm{W}^{l}}$ and refresh once per training step. Exploiting the rank-1 form of $\nabla_{\bm{W}^{l}}\ell_t$,
\begin{equation}
\big\langle \nabla_{\bm{W}^{l}}\ell_t,\, \bm{V}^{l}\big\rangle_F
\;=\;
(\bm{\delta}_t^{l})^{\!\top}\,\bm{V}^{l}\,\bm{h}_t^{l},
\label{eq:ap_gdp}
\end{equation}
which evaluates as a single matrix--vector product per layer per token without ever forming a model-sized intermediate. Stacking the activations of $T$ tokens into $\bm{H}^{l}\!\in\!\mathbb{R}^{T\times d_{\text{in}}}$ and the errors into $\bm{\Delta}^{l}\!\in\!\mathbb{R}^{T\times d_{\text{out}}}$, all $T$ scores collapse to one GEMM $\bm{H}^{l}(\bm{V}^{l})^{\!\top}\!\in\!\mathbb{R}^{T\times d_{\text{out}}}$ followed by an element-wise product with $\bm{\Delta}^{l}$, costing $\mathcal{O}(T d_{\text{in}} d_{\text{out}})$---identical in order to the activation--activation case and far below the $\mathcal{O}(T|\bm{\theta}|)$ baseline. The direct proxy term follows Eq.~\eqref{eq:ap_gdp}, while replacing $\bm{\delta}_t^{l}$ and $\bm{h}_t^{l}$ with $\alpha_{t\to k}^{l}\bm{\delta}_k^{l}$ and $\bm{h}_t^{l-1}$ gives $\Phi_{\text{prx}}^{\text{cau}}(y_t)$ analogously to Eq.~\eqref{eq:phi_tgt_causal} (see App.~\ref{app:gdp}).

\textbf{Total token valuation.}
Substituting the path-aware gradient $\nabla_\theta \ell_t^{\text{tot}}$ and the retention proxy $g_{\text{prx}}$ into the
valuation template Eq.~\eqref{eq:grad_alignment}, we obtain the AlphaToken
valuation:
\begin{equation}
\resizebox{0.86\linewidth}{!}{$
\begin{aligned}
\Phi(y_t)=\Phi_{\text{tgt}}^{\text{dir}}(y_t)+ \Phi_{\text{tgt}}^{\text{cau}}(y_t)+\lambda \big[\Phi_{\text{prx}}^{\text{dir}}(y_t)+ \Phi_{\text{prx}}^{\text{cau}}(y_t)\big].
\end{aligned}
$}
\label{eq:phi_total}
\end{equation}

By Eq.~\eqref{eq:phi_proxy}, the retention bracket is $\langle\nabla_{\bm{\theta}}\ell_t^{\text{tot}},\bm{g}_{\text{prx}}\rangle$. All four scores reuse one cached forward/backward pass; the extra cost is evaluating the fixed validation batch in the joint forward/backward pass to obtain
\((\bm{\delta}_{v}^{l},\bm{h}_{v}^{l})_{v\in\mathcal{V}_{\mathrm{tgt}}}\),
plus one GEMM per linear layer for A--P contraction with \(\bm{V}^{l}\).

\begin{algorithm}[t]
\caption{Value-Aware Post-Tuning}
\label{alg:value_aware_ft}
\begin{algorithmic}[1]
\Require Dataset $\mathcal{D}^{\mathrm{train}}_{\mathrm{tgt}}$, pre-trained $\bm{\theta}_{\text{ref}}$, Fisher $\bm{F}_{\text{ref}}$, validation set $\mathcal{D}_{\text{tgt}}^{\text{val}}$, ratio $\rho$, layer set $\mathcal{S}$, learning rate $\eta$, weight $\lambda$, window $W$.
\Ensure Optimized parameters $\bm{\theta}^\star$.
\For{each training step}
    \State Sample minibatch $\mathcal{B}\sim\mathcal{D}^{\mathrm{train}}_{\mathrm{tgt}}$.
    \State Compute $\Phi_{\text{tgt}}^{\text{dir/cau}}$ by Eqs.~\eqref{eq:phi_tgt_dir}, \eqref{eq:phi_tgt_causal}.
    \State Refresh $\bm{V}^{l}\!=\!\bm{F}_{\bm{W}^{l}}\!\odot\!(\bm{W}^{l}\!-\!\bm{W}^{l}_{\text{ref}})$.
    \State Compute $\Phi_{\text{prx}}^{\text{dir/cau}}$ via Eq.~\eqref{eq:ap_gdp}.
    \State Aggregate $\Phi(y_t)$ via Eq.~\eqref{eq:phi_total}; mask $m_t\!=\!\mathbb{I}[\Phi(y_t)\!\ge\!\tau_\rho]$ at within-batch top-$\rho$.
    \State Form $\mathcal{L}\!=\!\mathcal{L}_{\text{Alpha-SFT}}$ by Eq.~\eqref{eq:sft_masked_loss} or $\mathcal{L}_{\text{Alpha-DPO}}$ by
    Eq.~\eqref{eq:dpo_masked_loss}.
    \State $\bm{\theta}\leftarrow\bm{\theta}-\eta\nabla_{\bm{\theta}}\mathcal{L}$.
\EndFor
\State \textbf{return}\ $\bm{\theta}^\star$.
\end{algorithmic}
\end{algorithm}

\subsection{Value-Aware Post-Training}
\label{sec:value_aware_ft}
Given the per-token value $\Phi(y_t)$, AlphaToken allocates gradient flow to the most informative response tokens via binary masking.

\textbf{Supervised Fine-tuning (SFT).}
With a within-batch top-$\rho$ threshold $\tau_\rho$, the value-aware loss is
\begin{equation}
\resizebox{0.75\linewidth}{!}{$
\displaystyle
\mathcal{L}_{\text{Alpha-SFT}}(\bm{\theta})
= \sum_{t=1}^{T} \mathbb{I}\!\left[\Phi(y_t)\!\ge\!\tau_\rho\right] \ell_t(\bm{\theta}),
$}
\label{eq:sft_masked_loss}
\end{equation}
with $\ell_t(\bm{\theta})=-\!\log\pi_{\bm{\theta}}(y_t|\bm{x},\bm{y}_{<t})$. Low-utility tokens are skipped; gradient compute is reallocated to high-value positions.

\textbf{Direct preference optimization~(DPO).}
Extending the mask to DPO requires unrolling the sequence-level loss to per-token contributions. Given triplet $(\bm{x},\bm{y}^+,\bm{y}^-)$, with $s$ and $\omega\!\triangleq\!\beta\sigma(-s)$ as in standard DPO~\cite{rafailov2023direct},
\[
\resizebox{0.99\linewidth}{!}{$
\displaystyle
\nabla_{\bm{\theta}}\ell^{\text{DPO}}
=
\omega\big(\nabla_{\bm{\theta}}[-\!\log\pi_{\bm{\theta}}(\bm{y}^+|\bm{x})]
-\nabla_{\bm{\theta}}[-\!\log\pi_{\bm{\theta}}(\bm{y}^-|\bm{x})]\big),
$}
\]
so the layer-wise error signal scales linearly as $\bm{\delta}_t^{(l)\pm}=\pm\omega\,\tilde{\bm{\delta}}_t^{(l)\pm}$, where $\tilde{\bm{\delta}}_t^{l}$ is the standard SFT-style error signal. We then compute $\Phi(y_t^{\pm})$ via Eq.~\eqref{eq:phi_total} using the scaled $\bm{\delta}_t^{(l)\pm}$, and gate gradient flow with detached masks $m_t^{\pm}\!=\!\mathrm{sg}(\mathbb{I}[\Phi(y_t^{\pm})\!\ge\!\tau_\rho^{\pm}])$ and detached weight $\omega_{\mathrm{sg}}\!=\!\mathrm{sg}(\omega)$, where $\mathrm{sg}(\cdot)$ denotes the stop-gradient operator:
\begin{equation}
\resizebox{0.87\linewidth}{!}{$
\displaystyle
\mathcal{L}_{\text{Alpha-DPO}}(\bm{\theta})
=
\mathbb{E}\!\left[ \omega_{\mathrm{sg}}\!\sum_t\!\big( m_t^+\ell_t^+(\bm{\theta}) - m_t^-\ell_t^-(\bm{\theta}) \big) \right],
$}
\label{eq:dpo_masked_loss}
\end{equation}
with $\ell_t^{\pm}(\bm{\theta})\!=\!-\!\log\pi_{\bm{\theta}}(y_t^{\pm}|\bm{x},\bm{y}_{<t}^{\pm})$. The sequence-level DPO coefficient is computed from the unmasked preference logit and detached, so masking applies only to the per-token surrogate update. High-value tokens in $\bm{y}^+$ are reinforced, while selected tokens in $\bm{y}^-$ provide focused negative evidence whose likelihood is reduced, mitigating alignment tax without an explicit regularizer. Algorithm~\ref{alg:value_aware_ft} summarizes the unified procedure.

\newtcbox{\venuebox}{on line,
  boxrule=0pt,
  colback=red!15,
  colframe=gray!15,
  arc=1.5pt,
  left=2pt,right=2pt,top=1pt,bottom=1pt,
  boxsep=0pt}
  
\newcommand{\venuetag}[1]{{\scriptsize\venuebox{\textcolor{black!70}{#1}}}}

\begin{table*}[t]
\centering
\caption{Fine-Tuning main results. Pre-trained models are excluded from color ranking. The best and second-best results are highlighted in \textcolor{red}{red} and \textcolor{blue}{blue}, respectively.}
\vspace{-0.1in}
\label{tab:main_rq1_sft}
\scriptsize
\renewcommand{\arraystretch}{1.0}
\setlength{\tabcolsep}{6pt}
\definecolor{highlightpurple}{HTML}{E6E6FA}
\begin{tabular}{c|c|ccccc|c|c}
\toprule
\multirow{2}{*}[-0.6ex]{\textbf{Model}} &
\multirow{2}{*}[-0.6ex]{\textbf{Method}} &
\multicolumn{5}{c|}{\textbf{General Capability Acc.~(\%)}} &
\textbf{Target~(\%)} &
\textbf{Overall} \\
\cmidrule(lr){3-7}\cmidrule(lr){8-8}\cmidrule(lr){9-9}
& & \textbf{ARC-C} & \textbf{HellaSwag} & \textbf{MMLU} & \textbf{GSM8K} & \textbf{Avg.} & \textbf{HE} & \textbf{Avg.} \\
\midrule
& Pre-trained    & 43.00$_{\pm 0.00}$ & 55.82$_{\pm 0.00}$ & 56.54$_{\pm 0.00}$ & 27.45$_{\pm 0.00}$ & 45.70$_{\pm 0.00}$ & 28.66$_{\pm 0.00}$ & 37.18$_{\pm 0.00}$ \\
& Standard FT   & 39.97$_{\pm 0.47}$ & 52.45$_{\pm 0.34}$ & 48.23$_{\pm 0.43}$ & 20.58$_{\pm 0.34}$ & 40.31$_{\pm 0.20}$ & \textcolor{red}{44.60$_{\pm 0.48}$} & 42.46$_{\pm 0.26}$ \\
& LoRA   \venuetag{ICLR 2022}       & 41.62$_{\pm 0.39}$ & 53.32$_{\pm 0.34}$ & 50.42$_{\pm 0.35}$ & 21.00$_{\pm 0.39}$ & 41.59$_{\pm 0.18}$ & 41.46$_{\pm 0.40}$ & 41.53$_{\pm 0.22}$ \\
& LESS  \venuetag{ICML 2024}         & \textcolor{blue}{42.62$_{\pm 0.34}$} & \textcolor{blue}{54.62$_{\pm 0.30}$} & 53.57$_{\pm 0.31}$ & 22.61$_{\pm 0.23}$ & 43.36$_{\pm 0.15}$ & 42.24$_{\pm 0.34}$ & 42.80$_{\pm 0.19}$ \\
& Token Cleaning \venuetag{ICML 2025} & 41.77$_{\pm 0.37}$ & 52.20$_{\pm 0.21}$ & 53.94$_{\pm 0.33}$ & 26.00$_{\pm 0.24}$ & 43.48$_{\pm 0.15}$ & 40.93$_{\pm 0.36}$ & 42.21$_{\pm 0.19}$ \\
& STM  \venuetag{NeurIPS 2025}          & 42.02$_{\pm 0.35}$ & 54.61$_{\pm 0.34}$ & \textcolor{blue}{55.30$_{\pm 0.30}$} & \textcolor{blue}{26.22$_{\pm 0.34}$} & \textcolor{blue}{44.54$_{\pm 0.17}$} & 41.54$_{\pm 0.35}$ & 43.04$_{\pm 0.19}$ \\
& XTF     \venuetag{ICLR 2026}        & 40.47$_{\pm 0.41}$ & 52.03$_{\pm 0.29}$ & 50.69$_{\pm 0.36}$ & 22.58$_{\pm 0.29}$ & 41.44$_{\pm 0.17}$ & 42.02$_{\pm 0.39}$ & 41.73$_{\pm 0.21}$ \\
& ssTOKEN \venuetag{ICLR 2026}         & 41.94$_{\pm 0.37}$ & 53.37$_{\pm 0.31}$ & 53.10$_{\pm 0.33}$ & 25.38$_{\pm 0.32}$ & 43.45$_{\pm 0.17}$ & 42.93$_{\pm 0.38}$ & \textcolor{blue}{43.19$_{\pm 0.20}$} \\
\rowcolor{highlightpurple}\cellcolor{white}
\multirow{-9}{*}{\rotatebox{90}{\textbf{Llama-3.2-3B}}} & \textbf{AlphaToken} & \textcolor{red}{42.76$_{\pm 0.31}$} & \textcolor{red}{55.48$_{\pm 0.26}$} & \textcolor{red}{56.22$_{\pm 0.28}$} & \textcolor{red}{27.40$_{\pm 0.31}$} & \textcolor{red}{45.47$_{\pm 0.15}$} & \textcolor{blue}{43.98$_{\pm 0.40}$} & \textcolor{red}{44.73$_{\pm 0.21}$} \\
\midrule
& Pre-trained    & 51.45$_{\pm 0.00}$ & 56.86$_{\pm 0.00}$ & 59.60$_{\pm 0.00}$ & 37.00$_{\pm 0.00}$ & 51.23$_{\pm 0.00}$ & 35.36$_{\pm 0.00}$ & 43.30$_{\pm 0.00}$ \\
& Standard FT    & 50.17$_{\pm 0.45}$ & 52.63$_{\pm 0.38}$ & 53.87$_{\pm 0.40}$ & 24.12$_{\pm 0.28}$ & 45.20$_{\pm 0.19}$ & \textcolor{blue}{58.79$_{\pm 0.49}$} & 52.00$_{\pm 0.26}$ \\
& LoRA   \venuetag{ICLR 2022}       & 49.19$_{\pm 0.36}$ & 55.81$_{\pm 0.28}$ & 56.71$_{\pm 0.32}$ & 31.44$_{\pm 0.42}$ & 48.29$_{\pm 0.17}$ & 54.62$_{\pm 0.51}$ & 51.46$_{\pm 0.27}$ \\
& LESS  \venuetag{ICML 2024}         & 50.77$_{\pm 0.34}$ & \textcolor{red}{56.45$_{\pm 0.30}$} & 54.26$_{\pm 0.29}$ & 31.98$_{\pm 0.37}$ & 48.37$_{\pm 0.16}$ & 57.17$_{\pm 0.44}$ & 52.77$_{\pm 0.23}$ \\
& Token Cleaning \venuetag{ICML 2025} & \textcolor{blue}{50.85$_{\pm 0.38}$} & 55.84$_{\pm 0.23}$ & 55.94$_{\pm 0.34}$ & 33.00$_{\pm 0.38}$ & 48.91$_{\pm 0.17}$ & 57.12$_{\pm 0.44}$ & 53.02$_{\pm 0.23}$ \\
& STM  \venuetag{NeurIPS 2025}          & 50.30$_{\pm 0.40}$ & \textcolor{blue}{56.19$_{\pm 0.34}$} & 57.30$_{\pm 0.27}$ & \textcolor{blue}{36.21$_{\pm 0.38}$} & \textcolor{blue}{50.00$_{\pm 0.18}$} & 54.01$_{\pm 0.51}$ & 52.01$_{\pm 0.27}$ \\
& XTF     \venuetag{ICLR 2026}        & 48.70$_{\pm 0.39}$ & 53.88$_{\pm 0.35}$ & 56.33$_{\pm 0.33}$ & 34.44$_{\pm 0.39}$ & 48.34$_{\pm 0.18}$ & 56.64$_{\pm 0.52}$ & 52.49$_{\pm 0.28}$ \\
& ssTOKEN \venuetag{ICLR 2026}         & 50.58$_{\pm 0.37}$ & 54.31$_{\pm 0.31}$ & \textcolor{blue}{57.34$_{\pm 0.34}$} & 35.20$_{\pm 0.40}$ & 49.36$_{\pm 0.18}$ & 57.45$_{\pm 0.51}$ & \textcolor{blue}{53.41$_{\pm 0.27}$} \\
\rowcolor{highlightpurple}\cellcolor{white}
\multirow{-9}{*}{\rotatebox{90}{\textbf{Gemma-3-4B}}} & \textbf{AlphaToken} & \textcolor{red}{51.10$_{\pm 0.30}$} & 55.12$_{\pm 0.29}$ & \textcolor{red}{58.22$_{\pm 0.28}$} & \textcolor{red}{36.58$_{\pm 0.46}$} & \textcolor{red}{50.26$_{\pm 0.17}$} & \textcolor{red}{62.15$_{\pm 0.34}$} & \textcolor{red}{56.21$_{\pm 0.19}$} \\
\midrule
& Pre-trained    & 54.27$_{\pm 0.00}$ & 59.67$_{\pm 0.00}$ & 72.62$_{\pm 0.00}$ & 86.20$_{\pm 0.00}$ & 68.19$_{\pm 0.00}$ & 60.96$_{\pm 0.00}$ & 64.58$_{\pm 0.00}$ \\
& Standard FT    & 49.25$_{\pm 0.41}$ & 54.21$_{\pm 0.33}$ & 66.24$_{\pm 0.36}$ & 77.92$_{\pm 0.31}$ & 61.91$_{\pm 0.18}$ & \textcolor{blue}{75.94$_{\pm 0.41}$} & 68.93$_{\pm 0.22}$ \\
& LoRA   \venuetag{ICLR 2022}       & 51.27$_{\pm 0.33}$ & 56.40$_{\pm 0.32}$ & 69.60$_{\pm 0.29}$ & 81.97$_{\pm 0.24}$ & 64.81$_{\pm 0.15}$ & 70.20$_{\pm 0.24}$ & 67.51$_{\pm 0.14}$ \\
& LESS  \venuetag{ICML 2024}         & 50.24$_{\pm 0.30}$ & 57.11$_{\pm 0.31}$ & 69.68$_{\pm 0.26}$ & 81.23$_{\pm 0.37}$ & 64.57$_{\pm 0.16}$ & 72.82$_{\pm 0.35}$ & 68.70$_{\pm 0.19}$ \\
& Token Cleaning \venuetag{ICML 2025} & \textcolor{blue}{52.76$_{\pm 0.35}$} & 55.02$_{\pm 0.34}$ & 68.91$_{\pm 0.31}$ & \textcolor{blue}{83.54$_{\pm 0.44}$} & 65.06$_{\pm 0.18}$ & 73.58$_{\pm 0.32}$ & 69.32$_{\pm 0.18}$ \\
& STM  \venuetag{NeurIPS 2025}          & 52.65$_{\pm 0.28}$ & 57.43$_{\pm 0.32}$ & \textcolor{blue}{70.63$_{\pm 0.25}$} & 82.87$_{\pm 0.41}$ & 65.90$_{\pm 0.16}$ & 71.93$_{\pm 0.46}$ & 68.92$_{\pm 0.24}$ \\
& XTF     \venuetag{ICLR 2026}        & 52.56$_{\pm 0.36}$ & \textcolor{blue}{57.61$_{\pm 0.34}$} & 70.51$_{\pm 0.30}$ & \textcolor{red}{83.59$_{\pm 0.43}$} & \textcolor{blue}{66.07$_{\pm 0.18}$} & 73.61$_{\pm 0.41}$ & 69.84$_{\pm 0.22}$ \\
& ssTOKEN \venuetag{ICLR 2026}         & 51.39$_{\pm 0.37}$ & 55.17$_{\pm 0.35}$ & 70.52$_{\pm 0.31}$ & 82.72$_{\pm 0.42}$ & 64.95$_{\pm 0.18}$ & 75.21$_{\pm 0.41}$ & \textcolor{blue}{70.08$_{\pm 0.22}$} \\
\rowcolor{highlightpurple}\cellcolor{white}
\multirow{-9}{*}{\rotatebox{90}{\textbf{Qwen-3.5-9B}}} & \textbf{AlphaToken} & \textcolor{red}{53.13$_{\pm 0.27}$} & \textcolor{red}{57.71$_{\pm 0.31}$} & \textcolor{red}{71.20$_{\pm 0.25}$} & 83.14$_{\pm 0.39}$ & \textcolor{red}{66.30$_{\pm 0.15}$} & \textcolor{red}{78.88$_{\pm 0.34}$} & \textcolor{red}{72.59$_{\pm 0.19}$} \\
\bottomrule
\end{tabular}
\vspace{-0.20in}
\end{table*}

\section{Theoretical Analysis}
\label{sec:theory}
We justify the two approximations underlying AlphaToken: the Value-Propagation simplification of cross-position Jacobians and the data-free Fisher-drift proxy for retention stability. Complete bounds and proofs are in Appendix~\ref{app:theory}.

\paragraph{Value-Propagation approximation.}
The exact causal gradient requires back-propagating through the full softmax attention score matrix. Retaining only the dominant linear value-projection path incurs a controlled $\mathcal{O}(1/\sqrt{d_h})$ operator-norm error, $\| \bm{J}_{k,t}^{l} - \hat{\bm{J}}_{k,t}^{l} \|_{\mathrm{op}} \le (C/\sqrt{d_h})\,\alpha_{t\to k}^{l}\|\bm{v}_t - \bm{o}_k\|$, that vanishes under both sparse and saturated attention: small attention weights suppress the prefactor, while saturated attention forces the output to collapse to the value vector. See Appendix~\ref{app:value_path_bound}.

\paragraph{Data-free retention bound.}
The Fisher-drift proxy approximates the true retention gradient without any retention data. Their discrepancy is bounded by three terms, $\|\bm{r}_0\| + \|\bm{H}_{\text{ret}}\!-\!\bm{F}_{\text{ref}}\|\|\Delta\bm{\theta}\| + \tfrac{M_3}{2}\|\Delta\bm{\theta}\|^{2}$: \textbf{(I)}~a first-order residual at the reference checkpoint; \textbf{(II)}~a Fisher--Hessian mismatch decomposing into a calibration gap~\citep{martens2020new} and a Wasserstein distance~\citep{kunstner2019limitations}, both controllable without retention data; and \textbf{(III)}~a third-order remainder suppressed by value-aware masking because it limits parameter drift. None requires expectations over the unobserved retention dataset~$\mathcal{D}_{\text{ret}}$. See Appendix~\ref{app:retention_proxy_bound}.

\begin{table*}[t]
\centering
\caption{Preference Optimization main results. Base denotes the UltraChat-200k warm-started reference before preference optimization. The best and second-best results are highlighted in \textcolor{red}{red} and \textcolor{blue}{blue}, respectively.}
\vspace{-0.1in}
\label{tab:main_rq1_dpo}
\scriptsize
\renewcommand{\arraystretch}{1}
\setlength{\tabcolsep}{4pt}
\definecolor{highlightpurple}{HTML}{E6E6FA}
\begin{tabular}{l|ccccc|ccc|c}
\toprule
\multirow{2}{*}[-0.6ex]{\textbf{Method}} &
\multicolumn{5}{c|}{\textbf{General Capability Acc.~(\%)}} &
\multicolumn{3}{c|}{\textbf{Preference Win Rate~(\%)}} &
\textbf{Overall} \\
\cmidrule(lr){2-6}\cmidrule(lr){7-9}\cmidrule(lr){10-10}
& \textbf{ARC-C} & \textbf{HellaSwag} & \textbf{MMLU} & \textbf{GSM8K} & \textbf{Avg.}
& \textbf{AE2} & \textbf{A-Hard} & \textbf{Avg.}
& \textbf{Avg.} \\
\midrule
\rowcolor{black!6}\multicolumn{10}{c}{\textbf{Llama-3.2-3B}} \\
Base                         & 41.19$_{\pm 0.00}$ & 53.76$_{\pm 0.00}$ & 48.99$_{\pm 0.00}$ & 23.68$_{\pm 0.00}$ & 41.91$_{\pm 0.00}$ & 4.10$_{\pm 0.00}$ & 2.10$_{\pm 0.00}$ & 3.10$_{\pm 0.00}$ & 22.51$_{\pm 0.00}$ \\
DPO    \venuetag{NeurIPS 2023} & 39.84$_{\pm 0.52}$ & 52.20$_{\pm 0.50}$ & 47.62$_{\pm 0.38}$ & 21.05$_{\pm 0.68}$ & 40.18$_{\pm 0.49}$ & 12.86$_{\pm 0.38}$ & 9.40$_{\pm 0.47}$ & 11.13$_{\pm 0.30}$ & 25.66$_{\pm 0.29}$ \\
ConfPO \venuetag{ICML 2025}    & \textcolor{blue}{40.68$_{\pm 0.63}$} & 51.02$_{\pm 0.49}$ & 47.48$_{\pm 0.38}$ & \textcolor{blue}{22.65$_{\pm 0.74}$} & 40.46$_{\pm 0.50}$ & 16.83$_{\pm 0.49}$ & 13.30$_{\pm 0.36}$ & 15.07$_{\pm 0.30}$ & \textcolor{blue}{27.77$_{\pm 0.29}$} \\
SePO   \venuetag{EMNLP 2025}   & 37.15$_{\pm 0.50}$ & 50.48$_{\pm 0.48}$ & 46.01$_{\pm 0.40}$ & 20.72$_{\pm 0.70}$ & 38.59$_{\pm 0.50}$ & \textcolor{blue}{17.95$_{\pm 0.44}$} & \textcolor{blue}{13.90$_{\pm 0.41}$} & \textcolor{blue}{15.93$_{\pm 0.30}$} & 27.26$_{\pm 0.29}$ \\
TI-DPO \venuetag{ICLR 2026}    & 40.42$_{\pm 0.55}$ & \textcolor{blue}{52.77$_{\pm 0.51}$} & \textcolor{red}{48.25$_{\pm 0.39}$} & 22.26$_{\pm 0.72}$ & \textcolor{blue}{40.93$_{\pm 0.50}$} & 15.59$_{\pm 0.33}$ & 12.20$_{\pm 0.48}$ & 13.90$_{\pm 0.29}$ & 27.42$_{\pm 0.29}$ \\
\rowcolor{highlightpurple}
\textbf{AlphaToken} & \textcolor{red}{40.95$_{\pm 0.63}$} & \textcolor{red}{53.36$_{\pm 0.47}$} & \textcolor{blue}{48.03$_{\pm 0.39}$} & \textcolor{red}{23.15$_{\pm 0.71}$} & \textcolor{red}{41.37$_{\pm 0.49}$} & \textcolor{red}{19.44$_{\pm 0.47}$} & \textcolor{red}{15.80$_{\pm 0.39}$} & \textcolor{red}{17.62$_{\pm 0.31}$} & \textcolor{red}{29.50$_{\pm 0.29}$} \\
\midrule
\rowcolor{black!6}\multicolumn{10}{c}{\textbf{Gemma-3-4B}} \\
Base                         & 50.66$_{\pm 0.00}$ & 51.60$_{\pm 0.00}$ & 52.86$_{\pm 0.00}$ & 25.42$_{\pm 0.00}$ & 45.14$_{\pm 0.00}$ & 10.87$_{\pm 0.00}$ & 6.30$_{\pm 0.00}$ & 8.59$_{\pm 0.00}$ & 26.87$_{\pm 0.00}$ \\
DPO    \venuetag{NeurIPS 2023} & 49.42$_{\pm 0.51}$ & 50.58$_{\pm 0.50}$ & 51.35$_{\pm 0.40}$ & 22.86$_{\pm 0.77}$ & 43.55$_{\pm 0.51}$ & 22.48$_{\pm 0.41}$ & 19.40$_{\pm 0.35}$ & 20.94$_{\pm 0.27}$ & 32.25$_{\pm 0.29}$ \\
ConfPO \venuetag{ICML 2025}    & \textcolor{blue}{50.20$_{\pm 0.61}$} & \textcolor{blue}{51.28$_{\pm 0.51}$} & \textcolor{red}{52.34$_{\pm 0.40}$} & 23.38$_{\pm 0.82}$ & \textcolor{blue}{44.30$_{\pm 0.51}$} & \textcolor{blue}{28.45$_{\pm 0.48}$} & 24.90$_{\pm 0.37}$ & \textcolor{blue}{26.68$_{\pm 0.30}$} & \textcolor{blue}{35.49$_{\pm 0.30}$} \\
SePO   \venuetag{EMNLP 2025}   & 47.78$_{\pm 0.53}$ & 48.82$_{\pm 0.48}$ & 50.72$_{\pm 0.41}$ & 23.45$_{\pm 0.79}$ & 42.69$_{\pm 0.51}$ & 27.83$_{\pm 0.34}$ & \textcolor{blue}{25.40$_{\pm 0.49}$} & 26.62$_{\pm 0.30}$ & 34.66$_{\pm 0.30}$ \\
TI-DPO \venuetag{ICLR 2026}    & 50.02$_{\pm 0.49}$ & 51.05$_{\pm 0.49}$ & \textcolor{blue}{52.10$_{\pm 0.39}$} & \textcolor{blue}{23.96$_{\pm 0.80}$} & 44.28$_{\pm 0.51}$ & 26.77$_{\pm 0.45}$ & 23.20$_{\pm 0.31}$ & 24.99$_{\pm 0.27}$ & 34.64$_{\pm 0.29}$ \\
\rowcolor{highlightpurple}
\textbf{AlphaToken} & \textcolor{red}{50.48$_{\pm 0.61}$} & \textcolor{red}{51.46$_{\pm 0.47}$} & 51.68$_{\pm 0.39}$ & \textcolor{red}{24.06$_{\pm 0.81}$} & \textcolor{red}{44.42$_{\pm 0.51}$} & \textcolor{red}{31.86$_{\pm 0.49}$} & \textcolor{red}{27.40$_{\pm 0.41}$} & \textcolor{red}{29.63$_{\pm 0.32}$} & \textcolor{red}{37.03$_{\pm 0.30}$} \\
\midrule
\rowcolor{black!6}\multicolumn{10}{c}{\textbf{Qwen-3.5-9B}} \\
Base                         & 50.41$_{\pm 0.00}$ & 54.18$_{\pm 0.00}$ & 67.27$_{\pm 0.00}$ & 78.18$_{\pm 0.00}$ & 62.51$_{\pm 0.00}$ & 15.84$_{\pm 0.00}$ & 12.40$_{\pm 0.00}$ & 14.12$_{\pm 0.00}$ & 38.32$_{\pm 0.00}$ \\
DPO    \venuetag{NeurIPS 2023} & 49.02$_{\pm 0.52}$ & 52.84$_{\pm 0.49}$ & 65.71$_{\pm 0.36}$ & 75.62$_{\pm 0.89}$ & 60.80$_{\pm 0.45}$ & 34.41$_{\pm 0.42}$ & 29.80$_{\pm 0.36}$ & 32.11$_{\pm 0.28}$ & 46.46$_{\pm 0.27}$ \\
ConfPO \venuetag{ICML 2025}    & \textcolor{blue}{49.85$_{\pm 0.56}$} & \textcolor{blue}{53.72$_{\pm 0.49}$} & 66.02$_{\pm 0.34}$ & 76.36$_{\pm 0.93}$ & 61.49$_{\pm 0.46}$ & 35.84$_{\pm 0.47}$ & \textcolor{blue}{31.80$_{\pm 0.40}$} & 33.82$_{\pm 0.31}$ & \textcolor{blue}{47.66$_{\pm 0.28}$} \\
SePO   \venuetag{EMNLP 2025}   & 48.34$_{\pm 0.50}$ & 52.15$_{\pm 0.50}$ & 65.05$_{\pm 0.35}$ & 75.31$_{\pm 0.91}$ & 60.21$_{\pm 0.45}$ & \textcolor{blue}{36.46$_{\pm 0.39}$} & 31.60$_{\pm 0.48}$ & \textcolor{blue}{34.03$_{\pm 0.31}$} & 47.12$_{\pm 0.27}$ \\
TI-DPO \venuetag{ICLR 2026}    & 49.62$_{\pm 0.62}$ & 53.48$_{\pm 0.48}$ & \textcolor{blue}{66.38$_{\pm 0.36}$} & \textcolor{blue}{76.94$_{\pm 0.92}$} & \textcolor{blue}{61.61$_{\pm 0.45}$} & 34.16$_{\pm 0.33}$ & 30.50$_{\pm 0.45}$ & 32.33$_{\pm 0.28}$ & 46.97$_{\pm 0.26}$ \\
\rowcolor{highlightpurple}
\textbf{AlphaToken} & \textcolor{red}{50.18$_{\pm 0.54}$} & \textcolor{red}{54.02$_{\pm 0.48}$} & \textcolor{red}{67.05$_{\pm 0.35}$} & \textcolor{red}{77.92$_{\pm 0.90}$} & \textcolor{red}{62.29$_{\pm 0.45}$} & \textcolor{red}{38.76$_{\pm 0.49}$} & \textcolor{red}{34.60$_{\pm 0.37}$} & \textcolor{red}{36.68$_{\pm 0.31}$} & \textcolor{red}{49.49$_{\pm 0.27}$} \\
\bottomrule
\end{tabular}
\vspace{-0.20in}
\end{table*}

\section{Experiments}
\label{sec:exp}

We design a series of experiments to answer the following four questions:
\circnum{1} Does AlphaToken improve the trade-off between target adaptation and stability under both SFT and preference optimization?
\circnum{2} Do the estimated token values provide an informative ranking for token selection?
\circnum{3} How much does each factorized component contribute, including adaptation vs.\ stability and direct vs.\ causal paths?
\circnum{4} How sensitive is AlphaToken to its key design choices, and what computational overhead does it introduce?

\subsection{Setup}
\label{sec:exp:setup}

\textbf{Models and training pipelines.}
We evaluate three representative pre-trained backbones of different scales: Llama-3.2-3B~\citep{grattafiori2024llama3herdmodels}, Gemma-3-4B~\citep{gemmateam2025gemma3technicalreport}, and Qwen-3.5-9B~\citep{qwen3.5}. 
For SFT, each model is directly fine-tuned on the target corpus. 
For preference optimization, we first perform a uniform SFT warm-up on UltraChat-200k~\citep{ding2023enhancing} to obtain a chat-capable reference model, and then optimize on UltraFeedback~\citep{notus2023}. 
Implementation details are provided in App.~\ref{app:exp:setup}.

\textbf{Datasets \& Evaluation Metrics.}
For SFT, we use Magicoder~\citep{wei2024magicoder} as the target corpus and evaluate target-task adaptation on HumanEval~\citep{chen2021evaluating}. 
For preference optimization, we train on UltraFeedback and evaluate preference learning on AlpacaEval~2~\citep{alpaca_eval} and Arena-Hard~v0.1~\citep{arenahard2024}. 
Retention is measured by general-capability accuracy on ARC-C~\citep{clark2018think}, HellaSwag~\citep{zellers2019hellaswag}, MMLU~\citep{hendrycks2021measuring}, and GSM8K~\citep{cobbe2021training}. 
Overall denotes the average of the target-side and retention-side scores~\citep{sanyal2025upweighting}.

\textbf{Baselines \& Implementation Details.}
For SFT, we compare against Standard FT, LoRA~\citep{hu2022lora}, LESS~\citep{xia2024less}, Token Cleaning~\citep{pang2025token}, STM~\citep{wu2025mitigating}, XTF~\citep{yang2026explainable}, and ssTOKEN~\citep{qin2026sstoken}.
For PO, we compare against DPO~\citep{rafailov2023direct}, SePO~\citep{yang2025selective}, TI-DPO~\citep{yang2026tokenimportance}, and ConfPO~\citep{yoon2025confpo}.
All baselines are implemented under the matched backbone, data split, training budget, tokenizer, optimizer, scheduler, and evaluation protocol.
For AlphaToken, we by default set the selection ratio $\rho{=}0.5$, the stability weight $\lambda{=}1.5$, the causal window $W{=}32$, and use the last $K{=}3$ Transformer layers for GDP across all backbones. All results are averaged over three seeds, with stds reported where shown.
Full hyperparameters and baseline reproduction details are reported in App.~\ref{app:exp}.
All experiments are conducted on $4{\times}$NVIDIA A100 GPUs.

\textbf{Validation and Monte-Carlo Fisher construction.}
For each target training set $\mathcal{D}_{\mathrm{tgt}}$, we randomly hold out a 32-sample target-validation subset $\mathcal{D}^{\mathrm{val}}_{\mathrm{tgt}}$ prior to training. In addition, 1,000 prompts are sampled from $\mathcal{D}_{\mathrm{tgt}}$ to construct the diagonal Monte-Carlo Fisher $\mathbf{F}_{\mathrm{ref}}$; these prompts are excluded from training to avoid any parameter updates. The training pool is therefore
\[
\mathcal{D}^{\mathrm{train}}_{\mathrm{tgt}} = \mathcal{D}_{\mathrm{tgt}} \setminus \big(\mathcal{D}^{\mathrm{val}}_{\mathrm{tgt}} \cup \mathcal{D}^{\mathrm{Fisher}}_{\mathrm{ref}}\big),
\]
where $\mathcal{D}^{\mathrm{Fisher}}_{\mathrm{ref}}$ denotes the 1,000 prompts randomly sampled for Fisher matrix construction. The target-validation samples are used only to compute target-side gradients for token valuation and do not directly contribute to the masked training loss.
 
\subsection{Main Results}
\label{sec:exp:rq1}
\paragraph{Supervised fine-tuning.}
Table~\ref{tab:main_rq1_sft} reports SFT results. AlphaToken achieves the best Overall across three backbones, improving over the strongest baseline by $1.54$, $2.80$, and $2.51$ points on Llama-3.2-3B, Gemma-3-4B, and Qwen-3.5-9B, respectively. These gains do not come at the cost of target adaptation: AlphaToken ranks first on HumanEval for Gemma-3-4B and Qwen-3.5-9B and second on Llama-3.2-3B, while matching or exceeding the strongest competitor on General Capability Avg. Baselines such as Token Cleaning and STM rely on a local heuristic that entangles adaptation and retention. AlphaToken derives a principled valuation $\Phi(y_t)$ via gradient alignment with the validation objective, scoring adaptation and retention separately, allowing masking to improve both. See App.~\ref{app:more_datasets_baselines} for MetaMathQA results.

\textbf{Preference optimization.}
Table~\ref{tab:main_rq1_dpo} reports the DPO comparison after UltraChat-200k warm-up. AlphaToken attains the best Overall score on all three backbones, improving over the strongest competing baseline by $1.73$, $1.54$, and $1.83$ points. The improvement is two-sided: Preference Avg. rises by $2.55$, $2.95$, and $2.86$ points, while General Capability Avg. rises by $0.91$, $0.12$, and $0.80$ points. This contrasts with the standard DPO row, where preference-side gains incur visible retention regression, and with ConfPO, SePO, and TI-DPO, which mitigate the alignment tax only on one side. AlphaToken instead masks both branches by the composite valuation $\Phi(y_t)$, yielding the two-sided gain observed above.
\subsection{Token Ranking Effectiveness}
\label{sec:exp:rq2}

\begin{figure}[t]
\centering
\includegraphics[width=\linewidth]
{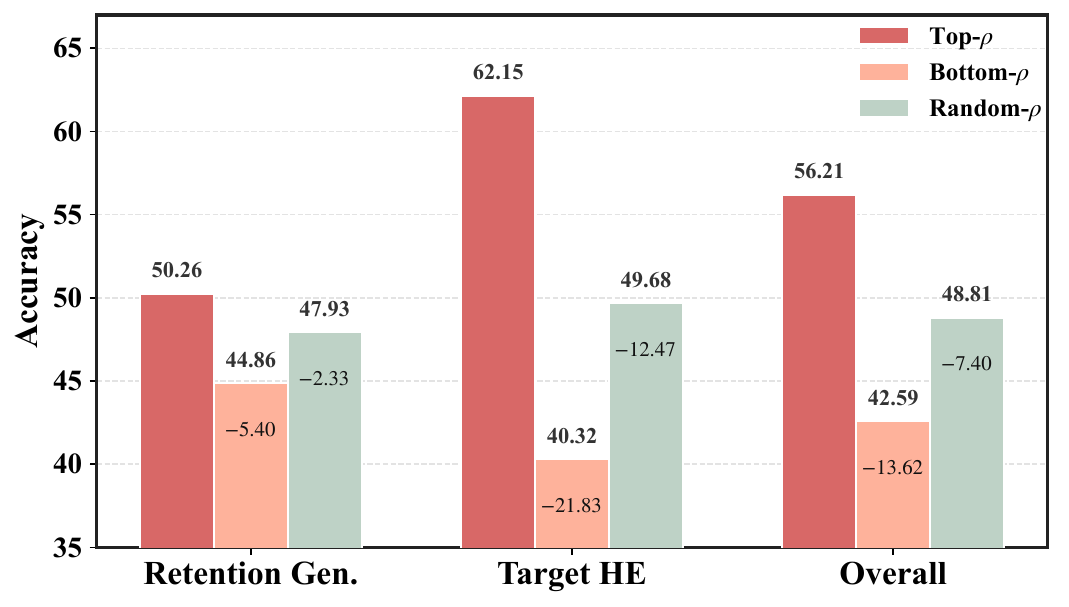}
\vspace{-0.3in}
\caption{
Token-policy intervention on Gemma-3-4B/SFT. 
All policies use the same $\rho{=}0.5$. 
}
\vspace{-0.15in}
\label{fig:intervention_rq2}
\vspace{-0.05in}
\end{figure}

We test whether the value score $\Phi$ induces a meaningful token ranking. 
On Gemma-3-4B/SFT, we compare the default Top-$\rho$ policy with two controls under the same retained-token ratio: Random-$\rho$, which selects tokens at random, and Bottom-$\rho$, which selects the lowest-scored tokens. 
As shown in Fig.~\ref{fig:intervention_rq2}, Top-$\rho$ substantially outperforms both controls. 
It improves the Overall score by $13.62$ points over Bottom-$\rho$ and by $7.40$ points over Random-$\rho$, with consistent gains in retention and target accuracy.
This clearly verifies that AlphaToken's score is not merely a masking heuristic: high-scored tokens are indeed more beneficial for training than low-scored or randomly selected tokens.


\subsection{Component Ablations} \label{sec:exp:rq3}
Table~\ref{tab:ablation_rq3} ablates  AlphaToken on Gemma-3-4B/SFT along two
axes: the objective axis (adaptation vs.\ stability) specifies what
a token's value rewards, and the path axis (direct vs.\ causal) specifies how
that value is attributed along the trajectory. The full design attains
$56.21$ Overall, above every single-axis variant.
On the objective axis, Adaptation-only gains $+3.33$ on Target HE but loses
$-3.62$ on Gen., while Stability-only shows the mirror pattern
($+1.92$/$-3.43$). The oppositely signed shifts indicate that the
two terms encode opposing update directions, toward the target vs.\ preserving
the pretraining distribution, so only their combination yields a balanced
point. On the path axis, Direct-only lowers Overall by $1.49$ with the loss
concentrated on Target HE, whereas Causal-only degrades it by
$3.27$, hurting both, indicating the direct signal is a low-variance backbone and
the causal signal supplies long-range credit assignment, and only their
combination recovers $56.21$.
\begin{table}[t]
\centering
\scriptsize
\caption{
Component ablations on Gemma-3-4B/SFT. 
AlphaToken decomposes token value along two axes: the objective axis, i.e., adaptation vs.\ stability, and the path axis, i.e., direct vs.\ causal.
}
\vspace{-0.1in}
\label{tab:ablation_rq3}
\setlength{\tabcolsep}{2.8pt}
\renewcommand{\arraystretch}{1.05}
\resizebox{\linewidth}{!}{
\begin{tabular}{@{}l|l|cc|c@{}}
\toprule
\textbf{Ablation Axis} & \textbf{Variant} & \textbf{Gen.} & \textbf{Target HE} & \textbf{Overall} \\
\midrule
Full design & Full AlphaToken & 50.26 & 62.15 & \textbf{56.21} \\
\midrule
\multirow{2}{*}{Objective} 
& Adaptation-only & 46.64 & \textbf{65.48} & 56.06 \\
& Stability-only  & \textbf{52.18} & 58.72 & 55.45 \\
\midrule
\multirow{2}{*}{Path}
& Direct-only & 50.38 & 59.06 & 54.72 \\
& Causal-only & 49.54 & 56.34 & 52.94 \\
\bottomrule
\end{tabular}
}
\vspace{-0.05in}
\end{table}

\subsection{Sensitivity and Computational Cost}
\label{sec:exp:sensitivity}

\begin{figure}[t]
\centering
\includegraphics[width=\linewidth]{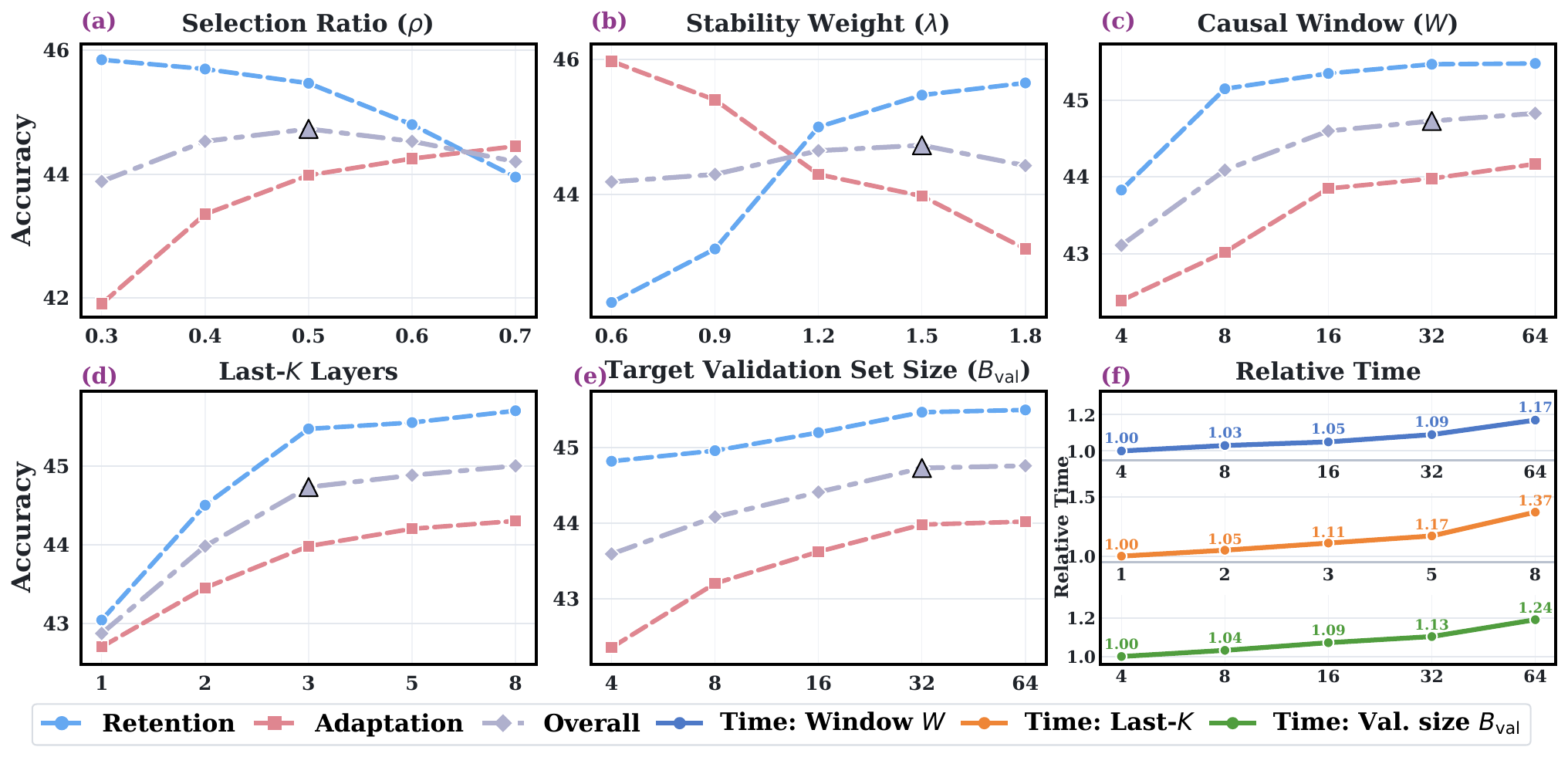}
\vspace{-0.3in}
\caption{Parameter sensitivity of AlphaToken on Llama-3.2-3B/SFT
and relative time along $W, K, B_{\mathrm{val}}$.}
\label{fig:sensitivity}
\vspace{-0.15in}
\end{figure}

Figure~\ref{fig:sensitivity} studies AlphaToken's key hyperparameters
$(\rho, \lambda, W, K, B_{\mathrm{val}})$ and per-step cost on Llama-3.2-3B/SFT.
For the selection ratio $\rho$, retention decreases and adaptation
increases monotonically as $\rho$ grows, and overall score peaks at
$\rho{=}0.5$.
The stability weight $\lambda$ trades retention against adaptation, with overall score peaking at
$\lambda{=}1.5$.
For the causal window, indicators plateau around $W{=}32$; further
enlargement still improves accuracy marginally but incurs additional
compute.
For the scoring depth, $K{=}3$ marks a clear knee beyond which gains
diminish while step time keeps growing roughly linearly in $K$.
For the target-validation batch, performance gains saturate at
$B_{\mathrm{val}}{=}32$.

\subsection{Component-wise Token Attribution}
\label{subsec:component_attribution}

\begin{figure}[t]
\centering
\includegraphics[width=\linewidth]{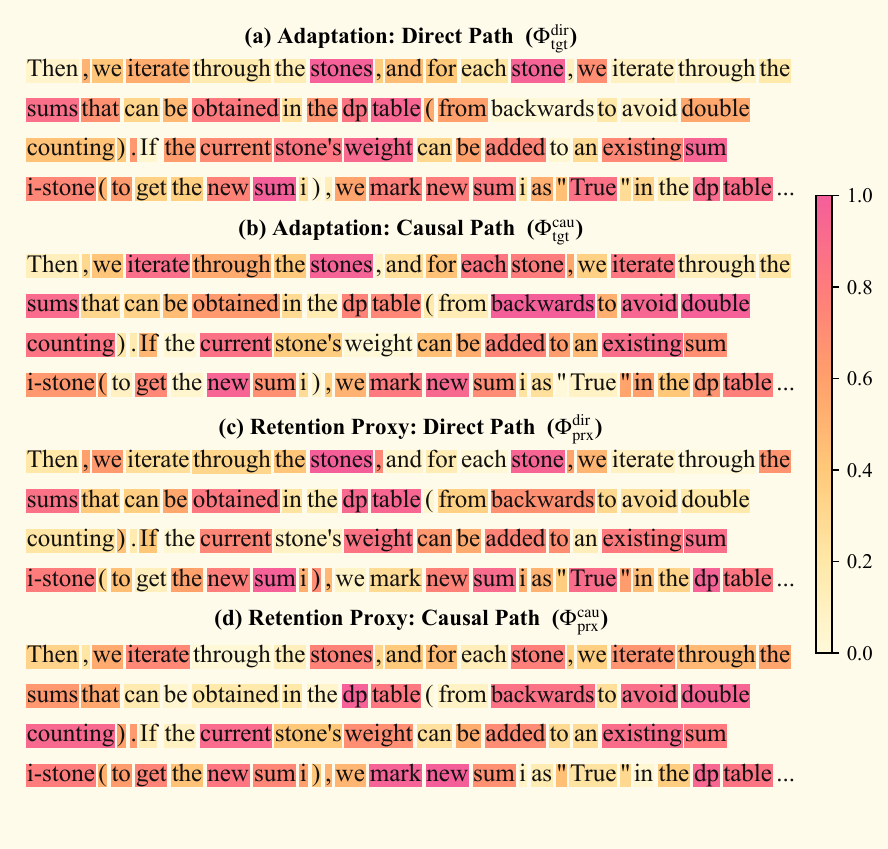}
\caption{
Component-wise token attribution on a dynamic-programming explanation. 
The same response segment is visualized under four components. 
}
\label{fig:component_attribution}
\vspace{-0.2in}
\end{figure}

We visualize the four AlphaToken scoring components on a representative dynamic-programming explanation from the \textit{Last Stone Weight II} task.
Scores are normalized within the same response segment, with darker colors indicating higher relative importance.
Figure~\ref{fig:component_attribution} shows that the components capture distinct token-level signals.
The adaptation direct path highlights task-relevant tokens such as \textit{stones}, \textit{sums}, \textit{dp table}, and \textit{weight}, while the adaptation causal path emphasizes reasoning-organizing tokens such as \textit{iterate}, \textit{backwards}, and \textit{avoid double counting}.
The retention-proxy components similarly focus on stable algorithmic concepts and structural reasoning tokens.
This case study suggests that the four components are not redundant and that direct and causal paths provide distinguishable valuation signals.

\paragraph{Additional robustness analyses.}
App.~\ref{app:fisher_proxy_analysis} analyzes the Fisher-drift proxy; App.~\ref{app:exp:value_path_fidelity} diagnoses the value-propagation approximation.
Apps.~\ref{app:exp:token_tsne}--\ref{app:exp:additional_token_heatmaps} provide token coverage, representation stability, and attribution analyses.
Apps.~\ref{app:exp:runtime}--\ref{app:exp:dpo_sensitivity} report efficiency, additional dataset and baselines, preference-optimization ablations, and sensitivity.

\section{Conclusion}
We introduce AlphaToken, a response-token valuation framework
for LLM post-training. AlphaToken decouples token value into two
objectives, target adaptation and retention stability, and splits
each objective along two paths by combining the direct 
signal with its downstream causal-path influence. As retention data
are typically unavailable in practice, a Fisher-drift proxy anchored
at the pre-trained reference replaces the inaccessible retention
gradient, and a Ghost Dot-Product family extended to
token-level computes all four scores at activation-space cost.
Experiments on SFT and PO across 3B to 9B backbones confirm that
AlphaToken improves the adaptation and retention
trade-off while mitigating catastrophic forgetting.

\section*{Limitations}
We discuss limitations.
(i) Computational overhead. Under default $K{=}3$, $W{=}32$,
and $B_{\mathrm{val}}{=}32$, per-step overhead remains modest. Nevertheless, the
scoring cost grows with the number of scored layers, causal window,
and validation batch size, so $K$, $W$, and $B_{\mathrm{val}}$ may
need re-budgeting for deeper models or longer contexts. (ii) Retention-proxy assumptions. The Fisher-drift proxy is tight
when the reference checkpoint is near-stationary for inaccessible
retention loss and the Fisher--Hessian mismatch is small. For
under-trained or heavily domain-specialized checkpoints, residual
and calibration gap may increase, making the proxy noisier. (iii) Hard masking. We instantiate AlphaToken with a binary
top-$\rho$ mask. Although this is simple and efficient, it discretizes continuous token values into hard keep/drop decisions.
Soft reweighting, value-conditioned learning rates, and schedules over
$\rho$ are left to future work. We do not foresee ethical risks beyond
those inherent to LLM adaptation.

\bibliography{custom}

\newpage
\appendix
\section{Token Valuation via In-Run Shapley from an Optimal Masking View}
\label{app:optimal_masking_to_shapley}

\paragraph{Setup: fine-tuning as optimal masking.}
We reframe token-level data selection as an optimal masking problem
that controls which token-level losses contribute gradients during
optimization, a perspective shared by selective training and pruning
schemes for LLM adaptation
\citep{simoulin2024memory,pang2025token,wu2025mitigating}. Let
$\mathcal{D}_{\text{train}}=\{(\bm{x}^{(i)},\bm{y}^{(i)})\}$ be the training
corpus with response $\bm{y}^{(i)}=[y^{(i)}_1,\dots,y^{(i)}_{T_i}]$. We
attach a binary mask $\bm{m}^{(i)}\in\{0,1\}^{T_i}$ indicating whether
token $y^{(i)}_t$ contributes to the gradient. The collection
$\mathbf{M}=\{\bm{m}^{(i)}\}$ instantiates token selection as a discrete
control variable. The goal is an optimal $\mathbf{M}^{*}$ minimizing a
validation loss $\mathcal{L}_{\text{val}}$ on a held-out set
$\mathcal{D}_{\text{val}}=\{(\bm{x}_0,\bm{y}_0)\}$:
\begin{align}
\min_{\mathbf{M}}\ \;& \mathcal{L}_{\text{val}}(\bm{x}_0,\bm{y}_0;\bm{\theta}^{*}), \label{eq:val_objective}\\
\text{s.t.}\ \;& \bm{\theta}^{*}\!=\!\arg\min_{\bm{\theta}}\!\!\sum_{i\in\mathcal{D}_{\text{train}}}\!\!\!\mathcal{L}\!\big(\bm{x}^{(i)},\bm{y}^{(i)}\!\odot\!\bm{m}^{(i)};\bm{\theta}\big). \label{eq:train_objective}
\end{align}
Here $\bm{y}^{(i)}\odot\bm{m}^{(i)}$ implies the loss is computed only on
unmasked positions, equivalently, masked tokens contribute zero
gradient---a selective backpropagation that reallocates compute toward
high-value signals \citep{simoulin2024memory}. This is a token-level
analogue of example reweighting and subset selection
\citep{paul2021deep,killamsetty2021grad}.

\paragraph{Token-player utility.}
For each example $(\bm{x}^{(i)},\bm{y}^{(i)})$ define token \emph{players}
$u=(i,t)$ with token-level loss $\ell_{u}(\bm{\theta})$ (e.g., teacher-forced
negative log-likelihood). Following the Data Shapley convention, we take
the to-be-maximised global utility to be the negative validation loss
of the trained model,
\begin{equation}
\label{eq:global_utility_def}
\mathcal{U}(\mathbf{M})\,:=\,-\,\mathcal{L}_{\mathrm{val}}\!\big(\bm{x}_0,\bm{y}_0;\bm{\theta}^{*}(\mathbf{M})\big).
\end{equation}
Exact token-level Shapley values under $\mathcal{U}$ are computationally
prohibitive because the per-sequence player count is large
\citep{ghorbani2019data}, motivating an in-run surrogate.

\paragraph{In-run local utility along a single trajectory.}
Following \citet{wang2025data}, we replace $\mathcal{U}(\mathbf{M})$
by a sum of \emph{local} utilities along one optimization trajectory. Let
$\bm{\theta}_s$ be the parameters at iteration $s$, $\mathcal{B}_s$ the player
set induced by the minibatch, and $S_s\subseteq\mathcal{B}_s$ the retained
subset. The masked one-step update is
\begin{equation}
\label{eq:masked_update}
\bm{\theta}_{s+1}(S_s)=\bm{\theta}_s-\eta_s\!\sum_{u\in S_s}\!\nabla_{\bm{\theta}}\ell_{u}(\bm{\theta}_s),
\end{equation}
and the local utility is the induced validation-loss decrease,
\begin{equation}
\label{eq:local_utility_def}
\Delta_{s}(S_s)\,:=\,\mathcal{L}_{\mathrm{val}}(\bm{\theta}_s)-\mathcal{L}_{\mathrm{val}}\!\big(\bm{\theta}_{s+1}(S_s)\big).
\end{equation}
The global effect of token retention is approximated by accumulating
$\Delta_{s}$ across steps.

\paragraph{Closed-form token value via first-order linearization.}
A first-order Taylor expansion of $\mathcal{L}_{\mathrm{val}}(\bm{\theta}_{s+1}(S_s))$
around $\bm{\theta}_s$ gives
\begin{equation}
\label{eq:first_order_delta}
\begin{aligned}
\Delta_{s}(S_s)
&\approx -\nabla_{\bm{\theta}}\mathcal{L}_{\mathrm{val}}(\bm{\theta}_s)^{\!\top}\!\big(\bm{\theta}_{s+1}(S_s)-\bm{\theta}_s\big)\\
&=\eta_s\!\sum_{u\in S_s}\!\big\langle\nabla_{\bm{\theta}}\mathcal{L}_{\mathrm{val}}(\bm{\theta}_s),\,\nabla_{\bm{\theta}}\ell_{u}(\bm{\theta}_s)\big\rangle.
\end{aligned}
\end{equation}
Since $\widehat{\Delta}_{s}(S_s)=\eta_s\!\sum_{u\in S_s}\!a_{u,s}$ is
additive in $S_s$, the Shapley value of player $u$ at step $s$
collapses to its individual contribution,
\begin{equation}
\label{eq:shapley_closed_form_step}
\phi_{u,s}=\eta_s\big\langle\nabla_{\bm{\theta}}\mathcal{L}_{\mathrm{val}}(\bm{\theta}_s),\,\nabla_{\bm{\theta}}\ell_{u}(\bm{\theta}_s)\big\rangle.
\end{equation}
Within a single training step $\eta_s$ is shared across tokens and
absorbed into the threshold $\tau_\rho$; we therefore drop it from the main-text $\Phi(y_t)$ formula. Accumulating across steps, when desired, would simply weight steps by $\eta_s$.

\paragraph{Alignment--stability decomposition.}
AlphaToken uses a composite validation objective
\begin{equation}
\label{eq:composite_val_obj}
\mathcal{L}_{\mathrm{val}}(\bm{\theta})=\mathcal{L}_{\mathrm{align}}(\bm{\theta})+\lambda\,\mathcal{L}_{\mathrm{stab}}(\bm{\theta}),
\end{equation}
with $\mathcal{L}_{\mathrm{align}}$ encoding target-task (or preference)
utility and $\mathcal{L}_{\mathrm{stab}}$ encoding retention/stability. By
inner-product linearity, Eq.~\eqref{eq:shapley_closed_form_step}
decomposes as
\begin{equation}
\label{eq:decomposition}
\begin{aligned}
\phi_{u,s}=\;& \eta_s\,\big\langle\nabla\mathcal{L}_{\mathrm{align}}(\bm{\theta}_s),\,\nabla\ell_{u}(\bm{\theta}_s)\big\rangle\\
&+\,\eta_s\lambda\,\big\langle\nabla\mathcal{L}_{\mathrm{stab}}(\bm{\theta}_s),\,\nabla\ell_{u}(\bm{\theta}_s)\big\rangle.
\end{aligned}
\end{equation}
When retention validation data are unavailable, $\mathcal{L}_{\mathrm{stab}}$
may be replaced by any differentiable proxy; the derivation only
requires $\nabla\mathcal{L}_{\mathrm{stab}}$ to be computable.

\paragraph{Autoregressive downstream influence.}
Replacing $\ell_{u}$ with an augmented term $\tilde{\ell}_{u}$ whose
gradient aggregates the position-$t$ direct contribution and the
propagation to later positions preserves additivity, so
Eq.~\eqref{eq:shapley_closed_form_step} generalises to
$\phi_{u,s}=\eta_s\big\langle\nabla\mathcal{L}_{\mathrm{val}},\,\nabla\tilde{\ell}_{u}\big\rangle$, unifying immediate learning benefit and downstream predictive influence within the same in-run Shapley framework.

\paragraph{Position relative to prior work.}
Eqs.~\eqref{eq:val_objective}--\eqref{eq:train_objective} expose
existing token-level methods as \emph{restricted families} of masks:
low-perplexity token learning constrains $\bm{m}^{(i)}$ via perplexity
profiles \citep{wu2025mitigating}; Token Cleaning uses token-wise loss
disparity \citep{pang2025token}; explainable filtering applies
novelty/reasoning priors \citep{yang2026explainable}; SePO, TI-DPO,
and ConfPO impose token masks or weights to modulate preference
gradients
\citep{christopoulou2024sparsepo,yang2026tokenimportance,yoon2025confpo}.
These designs each optimize a single proxy signal and are not derived
as objective-aligned solutions to $\mathcal{L}_{\mathrm{val}}$, leaving
systematic control of plasticity vs.\ stability ad hoc
\citep{kirkpatrick2017overcoming,li2017learning}. Our derivation supplies
the missing link.

\section{Theoretical Analysis}
\label{app:theory}

This appendix consolidates every derivation and proof deferred from the main text. We organise the material into four parts. App.~\ref{app:causal_decomp} sets up the causal gradient decomposition, derives the analytical form of the cross-position Jacobian, and proves universality across SFT and DPO. App.~\ref{app:gdp} establishes the three GDP factorizations. App.~\ref{app:value_path_bound} gives the complete proof of the Value-Propagation approximation bound. App.~\ref{app:retention_proxy_bound} gives the complete proof of the retention proxy bound, including a fine-grained decomposition of the Fisher--Hessian gap. Notation matches Sec.~\ref{sec:Methodology} and Sec.~\ref{sec:theory}.

\subsection{Causal Gradient Decomposition}
\label{app:causal_decomp}

\subsubsection{Notation}
A decoder-only Transformer is parameterised by $\bm{\theta}\!=\!\{\bm{\theta}^{l}\}_{l=1}^{L}$. For input $\bm{x}$ and length-$T$ response $\bm{y}$, the layer-$l$ hidden state at position $t$ is $\bm{h}_t^{l}\!\in\!\mathbb{R}^{d}$, with $\bm{h}_t^{(0)}$ the input embedding. The training objective factorizes as $\mathcal{L}(\bm{\theta})\!=\!\sum_{k=1}^{T}\ell_k(\bm{\theta})$. Causal masking enforces that $\bm{h}_k^{l}$ depends strictly on $\{\bm{h}_i^{l-1}\}_{i\le k}$, so any token $y_t$ enters $\mathcal{L}$ both \emph{directly} via $\ell_t$ and \emph{indirectly} via every $\ell_k$ with $k\!>\!t$.

\subsubsection{Direct and Causal Components}
Define $\nabla_{\bm{\theta}}\mathcal{L}_t^{\text{tot}}$ as the gradient component causally attributable to position $t$ through the immediate self-term and the single-hop cross-position transfer. Routing each future loss $\ell_k$ ($k\!>\!t$) backward through the layer-$l$ attention transfer $\bm{J}_{k,t}^{l}$ and isolating the self-term $k\!=\!t$ yields
\begin{equation}
\resizebox{0.8\linewidth}{!}{$
\displaystyle
\nabla_{\bm{\theta}}\ell_t^{\text{tot}}
=
\nabla_{\bm{\theta}}\ell_t
+
\sum_{k=t+1}^{T}\sum_{l=1}^{L}
\bm{\delta}_k^{l}\,\bm{J}_{k,t}^{l}\,\bm{\Theta}_t^{l-1}.
$}
\label{eq:app_causal_decomp}
\end{equation}
with $\bm{\delta}_k^{l}\!\triangleq\!\partial\ell_k/\partial\bm{h}_k^{l}\!\in\!\mathbb{R}^{1\times d}$, $\bm{J}_{k,t}^{l}\!\triangleq\!\partial\bm{h}_k^{l}/\partial\bm{h}_t^{l-1}$, and $\bm{\Theta}_t^{l-1}\!\triangleq\!\partial\bm{h}_t^{l-1}/\partial\bm{\theta}$. Eq.~\eqref{eq:app_causal_decomp} reproduces Eq.~\eqref{eq:causal_decomp}. Each additional cross-position hop introduces one further Jacobian factor $\bm{J}_{\cdot,\cdot}^{l}$, so an $m$-hop path decays as $O(d_h^{-m/2})$ in the score-propagation component (App.~\ref{app:value_path_bound}); the single-hop transfer is the dominant channel under saturated and sparse attention regimes.

\subsubsection{Cross-Position Jacobian}
At layer $l$, attention projects $\bm{h}^{l-1}$ via $\bm{W}_Q^{l},\bm{W}_K^{l},\bm{W}_V^{l}$, producing weights $\alpha_{k,i}^{l}\!=\!\mathrm{softmax}_i\!\big(\bm{q}_k^{l}(\bm{k}_i^{l})^{\!\top}\!/\!\sqrt{d_h}\big)$ and output $\bm{o}_k^{l}\!=\!\sum_{i\le k}\alpha_{k,i}^{l}\bm{v}_i^{l}$. Differentiating $\bm{o}_k^{l}$ w.r.t.\ $\bm{h}_t^{l-1}$ for $t\!<\!k$ separates two channels:
\begin{equation}
\resizebox{0.8\linewidth}{!}{$
\displaystyle
\frac{\partial\bm{o}_k^{l}}{\partial\bm{h}_t^{l-1}}
\!=\!
\underbrace{\alpha_{k,t}^{l}\bm{W}_V^{l}}_{\text{Value-Propagation}}
\!+\!
\underbrace{\sum_{i\le k}\!\frac{\partial\alpha_{k,i}^{l}}{\partial\bm{h}_t^{l-1}}\!\otimes\!\bm{v}_i^{l}}_{\text{Score-Propagation}}.
$}
\label{eq:app_two_paths}
\end{equation}
Applying $\partial\alpha_{k,i}^{l}/\partial e_{k,t}^{l}\!=\!\alpha_{k,i}^{l}(\delta_{it}\!-\!\alpha_{k,t}^{l})$ with $e_{k,t}^{l}\!=\!\bm{q}_k^{l}(\bm{k}_t^{l})^{\!\top}/\sqrt{d_h}$ collapses the Score-Propagation to
\begin{equation}
\resizebox{0.8\linewidth}{!}{$
\displaystyle
\text{Path}_{\text{Score}}
\!=\!
\frac{\alpha_{k,t}^{l}}{\sqrt{d_h}}\!\big(\bm{v}_t^{l}\!-\!\bm{o}_k^{l}\big)\!\otimes\!\big(\bm{q}_k^{l}(\bm{W}_K^{l})^{\!\top}\big).
$}
\label{eq:app_score_path}
\end{equation}
Combining,
\begin{equation}
\resizebox{0.8\linewidth}{!}{$
\displaystyle
\bm{J}_{k,t}^{l}
\!=\!
\alpha_{k,t}^{l}\bm{W}_V^{l}
\!+\!
\frac{\alpha_{k,t}^{l}}{\sqrt{d_h}}\big(\bm{v}_t^{l}\!-\!\bm{o}_k^{l}\big)\!\otimes\!\big(\bm{q}_k^{l}(\bm{W}_K^{l})^{\!\top}\big),
$}
\label{eq:app_exact_jacobian}
\end{equation}
where the Score-Propagation inherits a $1/\sqrt{d_h}$ attenuation directly from dot-product attention, justifying the Value-Propagation approximation in App.~\ref{app:value_path_bound}.

\subsubsection{Universality across SFT and DPO}
\label{app:universality_decomposition}
\label{app:dpo_decomp}

Eq.~\eqref{eq:app_causal_decomp} is independent of the choice of $\ell_k$.

\paragraph{SFT.}
With $\mathcal{L}_{\text{SFT}}\!=\!\sum_{k}\ell_k$ and $\ell_k\!=\!-\!\log\pi_{\bm{\theta}}(y_k|\bm{x},\bm{y}_{<k})$, summing Eq.~\eqref{eq:app_causal_decomp} over $t$ trivially recovers $\nabla_{\bm{\theta}}\mathcal{L}_{\text{SFT}}\!=\!\sum_t\nabla_{\bm{\theta}}\ell_t^{\text{tot}}$.

\paragraph{DPO.}
The sequence-level objective $\mathcal{L}_{\text{DPO}}\!=\!-\!\log\sigma(s)$ unrolls via $\nabla_s[-\!\log\sigma(s)]\!=\!-\sigma(-s)$ into per-position pseudo-losses $\ell_k^{\pm}\!=\!\pm\omega(-\!\log\pi_{\bm{\theta}}(y_k^{\pm}|\bm{x},\bm{y}_{<k}^{\pm}))$ with $\omega\!\triangleq\!\beta\sigma(-s)$. Define the formal token-routing notation
\begin{equation}
\nabla_{\bm{\theta}}\ell_k^{+}\big|_{\text{via }t}
\;\triangleq\;
\sum_{l=1}^{L}\bm{\delta}_k^{l}\,\bm{J}_{k,t}^{l}\,\bm{\Theta}_t^{l-1},
\label{eq:app_via_t}
\end{equation}
i.e., the contribution of $\ell_k^{+}$ that backpropagates through position $t$'s hidden state at layer $l$. In a causal LM, $\nabla_{\bm{\theta}}\ell_k^{+}\!=\!\sum_{t\le k}\nabla_{\bm{\theta}}\ell_k^{+}|_{\text{via }t}$. Substituting and swapping the order of summation,
\begin{equation}
\resizebox{0.8\linewidth}{!}{$
\displaystyle
\sum_{k=1}^{|\bm{y}^+|}\!\sum_{t=1}^{k}\!\nabla_{\bm{\theta}}\ell_k^{+}|_{\text{via }t}
\!=\!
\sum_{t=1}^{|\bm{y}^+|}\!\Big(\nabla_{\bm{\theta}}\ell_t^{+}|_{\text{via }t}\!+\!\!\sum_{k=t+1}^{|\bm{y}^+|}\!\nabla_{\bm{\theta}}\ell_k^{+}|_{\text{via }t}\Big).
$}
\label{eq:app_dpo_unroll}
\end{equation}
The summand inside the parenthesis matches Eq.~\eqref{eq:app_causal_decomp} term-for-term; the rejected branch follows symmetrically with the sign of $\omega$ reversed.
\subsection{Ghost Dot-Product Factorizations}
\label{app:gdp}
\label{app:ghost_dot_product}

We prove the three GDP identities used in Sec.~\ref{sec:efficient}. Two elementary facts: every token-level linear-layer gradient is rank-$1$,
\begin{equation}
\nabla_{\bm{W}^{l}}\ell_t \;=\; \bm{\delta}_t^{l}\!\otimes\!\bm{h}_t^{l};
\label{eq:app_rank1}
\end{equation}
and the Frobenius inner product of two rank-1 tensors decouples,
\begin{equation}
\langle\bm{u}\!\otimes\!\bm{v},\,\bm{u}'\!\otimes\!\bm{v}'\rangle_F\;=\;\langle\bm{u},\bm{u}'\rangle\langle\bm{v},\bm{v}'\rangle.
\label{eq:app_rank1_factor}
\end{equation}
We never materialize full parameter gradients in any of the three forms below.

\subsubsection{Direct Token Alignment (A--A)}
For each validation token $v\!\in\!\mathcal{V}_{\text{tgt}}$, the per-token gradient $\nabla_{\bm{W}^{l}}\ell_{v}\!=\!\bm{\delta}_{v}^{l}\!\otimes\!\bm{h}_{v}^{l}$ is rank-1, so applying Eq.~\eqref{eq:app_rank1_factor} to Eq.~\eqref{eq:app_rank1} gives
\begin{equation}
\resizebox{0.8\linewidth}{!}{$
\displaystyle
\big\langle\nabla_{\bm{W}^{l}}\ell_t,\nabla_{\bm{W}^{l}}\ell_{v}\big\rangle_F
\!=\!
\big\langle\bm{\delta}_t^{l},\bm{\delta}_{v}^{l}\big\rangle\!\big\langle\bm{h}_t^{l},\bm{h}_{v}^{l}\big\rangle.
$}
\label{eq:app_aa_direct}
\end{equation}
Averaging over $\mathcal{V}_{\text{tgt}}$ keeps the score scale invariant to validation batch size; summing further over $l$ yields $\Phi_{\text{tgt}}^{\text{dir}}(y_t)$. This A--A factorization follows the construction of~\citet{wang2025data}.

\subsubsection{Causal Token Alignment (A--A causal)}
Under the Value-Propagation approximation $\hat{\bm{J}}_{k,t}^{l}\!=\!\alpha_{t\to k}^{l}\bm{W}_V^{l}$, where $\alpha_{t\to k}^{l}\!\equiv\!\alpha_{k,t}^{l}$, the cross-position contribution becomes rank-1 in the value-projection parameters,
\begin{equation}
\nabla_{\bm{W}_V^{l}}\ell_{t\to k}\;\approx\;\alpha_{t\to k}^{l}\big(\bm{\delta}_k^{l}\!\otimes\!\bm{h}_t^{l-1}\big),
\label{eq:app_causal_rank1}
\end{equation}
with the shorthand $\alpha_{t\to k}^{l}\!\equiv\!\alpha_{k,t}^{l}$. For each validation token \(v\in\mathcal{V}_{\mathrm{tgt}}\), applying
Eq.~\eqref{eq:app_rank1_factor} once more gives
\begin{equation}
\resizebox{0.8\linewidth}{!}{$
\displaystyle
\big\langle
\nabla_{\bm{W}_V^{l}}\ell_{t\to k},
\nabla_{\bm{W}_V^{l}}\ell_{v}
\big\rangle_F
\approx
\alpha_{t\to k}^{l}
\big\langle\bm{\delta}_k^{l},\bm{\delta}_{v}^{l}\big\rangle
\big\langle\bm{h}_t^{l-1},\bm{h}_{v}^{l-1}\big\rangle .
$}
\label{eq:app_aa_causal}
\end{equation}
Averaging this quantity over \(v\in\mathcal{V}_{\mathrm{tgt}}\) and summing
over \(l\in\mathcal{S}\) and future positions \(k\) yields
\(\Phi_{\mathrm{tgt}}^{\mathrm{cau}}(y_t)\).

\subsubsection{Activation--Parameter GDP for the Retention Proxy (A--P)}
The retention proxy $\bm{g}_{\text{prx}}\!=\!\bm{F}_{\text{ref}}(\bm{\theta}\!-\!\bm{\theta}_{\text{ref}})$ is a fixed parameter-space vector and is \emph{not} the back-propagated output of any sample-level loss; it cannot be expressed as $\bm{\delta}_{\text{val}}\!\otimes\!\bm{h}_{\text{val}}$, so Eq.~\eqref{eq:app_rank1_factor} does not apply. We exploit only the rank-1 form Eq.~\eqref{eq:app_rank1} of the token gradient. Writing $\bm{V}^{l}\!\triangleq\!\bm{F}_{\bm{W}^{l}}\!\odot\!(\bm{W}^{l}\!-\!\bm{W}_{\text{ref}}^{l})$, where $\bm{F}_{\bm{W}^{l}}\!\triangleq\!\mathrm{Diag}(\bm{F}_{\text{ref}})|_{\bm{W}^{l}}$ is the layer-$l$ slice of the diagonal Fisher (refreshed once per training step), and using the trace identity $\langle\bm{u}\!\otimes\!\bm{v},\bm{M}\rangle_F\!=\!\bm{u}^{\!\top}\!\bm{M}\bm{v}$,
\begin{equation}
\big\langle\nabla_{\bm{W}^{l}}\ell_t,\bm{V}^{l}\big\rangle_F
\;=\;
(\bm{\delta}_t^{l})^{\!\top}\,\bm{V}^{l}\,\bm{h}_t^{l}.
\label{eq:app_ap_gdp}
\end{equation}
Stacking activations for \(T\) tokens into
\(\bm{H}^{l}\in\mathbb{R}^{T\times d_{\mathrm{in}}}\) and errors into
\(\bm{\Delta}^{l}\in\mathbb{R}^{T\times d_{\mathrm{out}}}\), all \(T\) scores
collapse to one GEMM
\(\bm{H}^{l}(\bm{V}^{l})^{\top}\in\mathbb{R}^{T\times d_{\mathrm{out}}}\),
followed by an element-wise product with \(\bm{\Delta}^{l}\) and summation
over the output dimension, with complexity
\(\mathcal{O}(T d_{\mathrm{in}} d_{\mathrm{out}})\). We apply this factorization to all linear layers in attention and MLP blocks; embedding and lm\_head layers are excluded from the scoring layer set $\mathcal{S}$. For LoRA-adapted layers, $\bm{F}_{\bm{W}^{l}}$ is computed and contracted on the LoRA factors $A,B$ separately; the cache is small. The same construction with $\alpha_{t\to k}^{l}\bm{\delta}_k^{l}$ and $\bm{h}_t^{l-1}$ gives the causal proxy term analogously to Eq.~\eqref{eq:app_aa_causal}.

\subsection{Value-Propagation Approximation Bound}
\label{app:value_path_bound}

We prove the bound on the residual between the exact cross-position Jacobian and its Value-Propagation approximation $\hat{\bm{J}}_{k,t}^{l}\!=\!\alpha_{t\to k}^{l}\bm{W}_V^{l}$. From Eq.~\eqref{eq:app_exact_jacobian}, the residual equals the Score-Propagation component exactly:
\begin{equation}
\resizebox{0.8\linewidth}{!}{$
\displaystyle
\mathcal{R}_{\text{score}}
\!\triangleq\!
\bm{J}_{k,t}^{l}\!-\!\hat{\bm{J}}_{k,t}^{l}
\!=\!
\frac{\alpha_{t\to k}^{l}}{\sqrt{d_h}}\big(\bm{v}_t^{l}\!-\!\bm{o}_k^{l}\big)\!\otimes\!\big(\bm{q}_k^{l}(\bm{W}_K^{l})^{\!\top}\big)
$},
\label{eq:app_score_residual}
\end{equation}
where $\alpha_{t\to k}^{l}\!\equiv\!\alpha_{k,t}^{l}\!\in\![0,1]$ denotes the scalar attention weight by which future position $k$ attends to source position $t$.

\paragraph{Operator-norm bound.}
For any rank-1 outer product, $\|\bm{u}\!\otimes\!\bm{v}\|_{\text{op}}\!=\!\|\bm{u}\|_2\|\bm{v}\|_2$. Applying this to Eq.~\eqref{eq:app_score_residual} and bounding $\|\bm{q}_k^{l}(\bm{W}_K^{l})^{\!\top}\|_2\!\le\!\|\bm{q}_k^{l}\|_2\,\|\bm{W}_K^{l}\|_{\text{op}}$,
\begin{equation}
\resizebox{0.8\linewidth}{!}{$
\displaystyle
\|\mathcal{R}_{\text{score}}\|_{\text{op}}
\!\le\!
\frac{\alpha_{t\to k}^{l}}{\sqrt{d_h}}\,\|\bm{v}_t^{l}\!-\!\bm{o}_k^{l}\|_2\,\|\bm{q}_k^{l}\|_2\,\|\bm{W}_K^{l}\|_{\text{op}}.
$}
\end{equation}
Letting $Q\!\triangleq\!\sup_{k,l}\|\bm{q}_k^{l}\|_2$ and $C\!\triangleq\!Q\,\|\bm{W}_K^{l}\|_{\text{op}}$, both finite for any well-defined Transformer,
\begin{equation}
\boxed{\;\|\mathcal{R}_{\text{score}}\|_{\text{op}}\;\le\;\frac{C}{\sqrt{d_h}}\,\alpha_{t\to k}^{l}\,\big\|\bm{v}_t^{l}\!-\!\bm{o}_k^{l}\big\|_2\,.\;}
\label{eq:app_value_path_bound}
\end{equation}

\paragraph{Saturation analysis.}
Eq.~\eqref{eq:app_value_path_bound} vanishes at \emph{both} extremes of the attention regime:
\begin{itemize}[leftmargin=*,topsep=2pt,itemsep=1pt]
\item \emph{Sparse attention} ($\alpha_{t\to k}^{l}\!\to\!0$): the prefactor drives the residual to zero directly.
\item \emph{Saturated attention} ($\alpha_{t\to k}^{l}\!\to\!1$): the output $\bm{o}_k^{l}\!=\!\sum_i\alpha_{k,i}^{l}\bm{v}_i^{l}$ collapses to $\bm{v}_t^{l}$, so $\|\bm{v}_t^{l}\!-\!\bm{o}_k^{l}\|_2\!\to\!0$.
\end{itemize}
Combined with $\mathcal{O}(1/\sqrt{d_h})$, this proves $\mathcal{R}_{\text{score}}$ is small in both regimes. An empirical verification is given in App.~\ref{app:exp:value_path_fidelity}.

\subsection{Fisher Drift Proxy without Retention Data}
\label{app:retention_proxy_bound}

We prove the data-free retention proxy bound and decompose its Fisher--Hessian-gap term into individually controllable pieces. Let $\mathcal{L}_{\text{ret}}(\bm{\theta})\!\triangleq\!\mathbb{E}_{(\bm{x},\bm{y})\sim\mathcal{D}_{\text{ret}}}\![-\!\log p_{\bm{\theta}}(\bm{y}|\bm{x})]$ be the unobserved retention loss, and define
\begin{equation}
\bm{g}_{\text{ideal}}\!\triangleq\!\nabla_{\bm{\theta}}\mathcal{L}_{\text{ret}}(\bm{\theta}),
\quad
\bm{g}_{\text{prx}}\!\triangleq\!\bm{F}_{\text{ref}}(\bm{\theta}\!-\!\bm{\theta}_{\text{ref}}),
\end{equation}
with $\bm{F}_{\text{ref}}$ the Monte-Carlo Fisher of Sec.~\ref{eq:mc_fisher}. Our goal is a bound on $\|\bm{g}_{\text{ideal}}\!-\!\bm{g}_{\text{prx}}\|$ that never evaluates expectations on $\mathcal{D}_{\text{ret}}$.

\subsubsection{Second-Order Taylor Expansion}
Assume $\mathcal{L}_{\text{ret}}\!\in\!C^3$ on a neighbourhood containing both $\bm{\theta}_{\text{ref}}$ and $\bm{\theta}$, and let $\Delta\bm{\theta}\!\triangleq\!\bm{\theta}\!-\!\bm{\theta}_{\text{ref}}$. Taylor's theorem with mean-value remainder gives
\begin{equation}
\resizebox{0.8\linewidth}{!}{$
\displaystyle
\bm{g}_{\text{ideal}}
\!=\!
\bm{r}_0+\bm{H}_{\text{ret}}\Delta\bm{\theta}+\tfrac{1}{2}\nabla^{\!3}\mathcal{L}_{\text{ret}}(\bm{\xi})[\Delta\bm{\theta},\Delta\bm{\theta}],
$}
\label{eq:app_taylor}
\end{equation}
for some $\bm{\xi}\!\in\![\bm{\theta}_{\text{ref}},\bm{\theta}]$, where $\bm{r}_0\!\triangleq\!\nabla\mathcal{L}_{\text{ret}}(\bm{\theta}_{\text{ref}})$ and $\bm{H}_{\text{ret}}\!\triangleq\!\nabla^2\mathcal{L}_{\text{ret}}(\bm{\theta}_{\text{ref}})$. Subtracting $\bm{g}_{\text{prx}}\!=\!\bm{F}_{\text{ref}}\Delta\bm{\theta}$ and applying the triangle inequality,
\begin{equation}
\resizebox{0.8\linewidth}{!}{$
\displaystyle
\|\bm{g}_{\text{ideal}}\!-\!\bm{g}_{\text{prx}}\|
\!\le\!
\|\bm{r}_0\|+\|\bm{H}_{\text{ret}}\!-\!\bm{F}_{\text{ref}}\|\!\cdot\!\|\Delta\bm{\theta}\|+\tfrac{M_3}{2}\|\Delta\bm{\theta}\|^2,
$}
\label{eq:app_proxy_bound}
\end{equation}
where $M_3\!\triangleq\!\sup_{\bm{\xi}\in[\bm{\theta}_{\text{ref}},\bm{\theta}]}\|\nabla^{\!3}\mathcal{L}_{\text{ret}}(\bm{\xi})\|_{\text{op}}$.

\paragraph{Remark on $\|\bm{r}_0\|$.}
The retention loss is defined over a (potentially heterogeneous) collection of downstream tasks, and pre-training is performed on a different, language-modelling distribution. We therefore do \emph{not} claim $\|\bm{r}_0\|\!=\!0$; rather, $\|\bm{r}_0\|$ is small to the extent that retention performance inherits the pre-training optimum, and is upper-bounded empirically by measuring the gradient norm of the reference checkpoint on a held-out probe.

\subsubsection{Decomposing the Fisher--Hessian Gap}
As $\bm{F}_{\text{ref}}$ is diagonal by construction (Sec.~3.3), we decompose $\|\bm{H}_{\text{ret}}\!-\!\bm{F}_{\text{ref}}\|$ into a diagonalisation residual, a calibration gap, and a Fisher transfer error. We introduce the auxiliary \emph{diagonal retention Fisher}
\begin{equation}
\resizebox{0.86\linewidth}{!}{$
\displaystyle
\bm{F}_{\text{ret}}^{\dagger}\!\triangleq\!\mathrm{Diag}\!\Big(\mathbb{E}_{\bm{x}\sim\mathcal{D}_{\text{ret}},\,\tilde{\bm{y}}\sim p_{\bm{\theta}_{\text{ref}}}\!(\cdot|\bm{x})}\!\big[\nabla_{\bm{\theta}}\!\log p_{\bm{\theta}_{\text{ref}}}\!(\tilde{\bm{y}}|\bm{x})^{\odot 2}\big]\Big),
$}
\label{eq:app_F_ret_dagger}
\end{equation}
which uses retention prompts but model-sampled labels and serves purely as an analytical object. By the triangle inequality,
\begin{equation*}
\resizebox{0.99\linewidth}{!}{$
\displaystyle
\|\bm{H}_{\text{ret}}\!-\!\bm{F}_{\text{ref}}\|
\!\le\!
\underbrace{\|\bm{H}_{\text{ret}}\!-\!\mathrm{Diag}(\bm{H}_{\text{ret}})\|}_{\Delta_{\text{off}}}
\!+\!
\underbrace{\|\mathrm{Diag}(\bm{H}_{\text{ret}})\!-\!\bm{F}_{\text{ret}}^{\dagger}\|}_{\Delta_{\text{cal}}}
\!+\!
\underbrace{\|\bm{F}_{\text{ret}}^{\dagger}\!-\!\bm{F}_{\text{ref}}\|}_{\text{Fisher transfer}}.
$}
\label{eq:app_gap_decomp}
\end{equation*}

\paragraph{Diagonalisation residual $\Delta_{\text{off}}$.}
This term captures the off-diagonal mass of $\bm{H}_{\text{ret}}$ and reflects the standard diagonal-Fisher modeling choice adopted by EWC~\citep{kirkpatrick2017overcoming} and related methods. It depends only on the local curvature of $\mathcal{L}_{\text{ret}}$ at $\bm{\theta}_{\text{ref}}$ and is independent of the input distribution used to construct $\bm{F}_{\text{ref}}$. Since $\bm{g}_{\text{prx}}\!=\!\bm{F}_{\text{ref}}\Delta\bm{\theta}$ acts only through the diagonal subspace, $\Delta_{\text{off}}$ is absorbed multiplicatively by $\|\Delta\bm{\theta}\|$, which value-aware masking actively keeps small throughout training.

\paragraph{Calibration gap $\Delta_{\text{cal}}$.}
Under the maximum-likelihood condition---i.e., when $p_{\bm{\theta}_{\text{ref}}}(\bm{y}|\bm{x})$ matches the conditional law of $\mathcal{D}_{\text{ret}}$---the Fisher information equals the negative log-likelihood Hessian~\citep{martens2020new}, yielding $\bm{F}_{\text{ret}}^{\dagger}\!=\!\mathrm{Diag}(\bm{H}_{\text{ret}})$ and $\Delta_{\text{cal}}\!=\!0$. For a publicly released checkpoint, $\Delta_{\text{cal}}$ is upper-bounded by the residual KL divergence between the model's predictive distribution and the data-generating conditional, which is small for well-calibrated pre-trained models.

\paragraph{Fisher transfer error.}
Both $\bm{F}_{\text{ret}}^{\dagger}$ and $\bm{F}_{\text{ref}}$ evaluate at the same $\bm{\theta}_{\text{ref}}$ and use labels drawn from the same conditional $p_{\bm{\theta}_{\text{ref}}}(\cdot|\bm{x})$; they differ only in the input distribution. Letting $L_g$ be the Lipschitz constant of the map
\(
\bm{x}\!\mapsto\!\mathrm{Diag}\!\big(\mathbb{E}_{\tilde{\bm{y}}\sim p_{\bm{\theta}_{\text{ref}}}\!(\cdot|\bm{x})}\!\big[\nabla_{\bm{\theta}}\!\log p_{\bm{\theta}_{\text{ref}}}\!(\tilde{\bm{y}}|\bm{x})^{\odot 2}\big]\big)
\),
\begin{equation}
\resizebox{0.8\linewidth}{!}{$
\displaystyle
\|\bm{F}_{\text{ret}}^{\dagger}\!-\!\bm{F}_{\text{ref}}\|\;\le\;L_g\,\mathcal{W}_{\!2}\!\big(p_{\mathcal{D}_{\text{ret}}}\!(\bm{x}),\,p_{\mathcal{X}}\!(\bm{x})\big).
$}
\label{eq:app_wasser}
\end{equation}
This bound depends only on the linguistic representativeness of $\mathcal{X}$. Reusing $\mathcal{D}_{\text{tgt}}$ prompts as $\mathcal{X}$ keeps the Wasserstein distance small in practice, as both share the surface morphology of natural language.

\subsubsection{Final Bound and Data-Free Reliability}
Combining the residual, Fisher--Hessian mismatch, diagonalization, and higher-order terms, we obtain
\begin{equation*}
\resizebox{0.95\linewidth}{!}{$
\displaystyle
\|\bm{g}_{\text{ideal}}\!-\!\bm{g}_{\text{prx}}\|
\!\le\!
\|\bm{r}_0\|+(\Delta_{\text{off}}\!+\!\Delta_{\text{cal}}\!+\!L_g\mathcal{W}_2)\!\cdot\!\|\Delta\bm{\theta}\|+\tfrac{M_3}{2}\|\Delta\bm{\theta}\|^2.
$}
\label{eq:app_final_bound}
\end{equation*}
Each of $\{\|\bm{r}_0\|,\Delta_{\text{cal}},L_g\mathcal{W}_2,M_3\}$ is controlled under conditions our framework already enforces or that public checkpoints already satisfy, while the diagonalisation residual $\Delta_{\text{off}}$ follows the standard diagonal-Fisher modeling choice and is absorbed multiplicatively through $\|\Delta\bm{\theta}\|$, which value-aware masking keeps small throughout training. The decomposition introduces $\mathcal{D}_{\text{ret}}$ only as an analytical object for bounding the approximation error---AlphaToken never evaluates these expectations or accesses retention examples during training, and only uses $\bm{F}_{\text{ref}}$ constructed from $\mathcal{X}$.
\section{Experiment Details}
\label{app:exp}

\subsection{Baseline Details}
\label{app:exp:baselines}

We compare AlphaToken with representative baselines for supervised fine-tuning and preference optimization.
\paragraph{Supervised fine-tuning baselines.}
For SFT, we compare against full fine-tuning, parameter-efficient tuning, sample selection, token selection, and forgetting-resistant methods.

\begin{itemize}[leftmargin=*]
    \item \textbf{Pre-trained}: The original backbone before target-task fine-tuning, used as the reference point for measuring adaptation and forgetting.

    \item \textbf{Standard FT}: Full-parameter fine-tuning with the standard response-token cross-entropy loss.

    \item \textbf{LoRA}~\citep{hu2022lora}: A parameter-efficient baseline that freezes the backbone and updates low-rank adapter parameters.

    \item \textbf{LESS}~\citep{xia2024less}: A sample-level selection baseline that selects training examples by gradient similarity to a target validation set.

    \item \textbf{Token Cleaning}~\citep{pang2025token}: A token-filtering baseline that masks low-quality tokens using token-wise loss-discrepancy signals.

    \item \textbf{STM}~\citep{wu2025mitigating}: A token-masking baseline that selects response tokens according to local likelihood or perplexity-based statistics to mitigate forgetting.

    \item \textbf{XTF}~\citep{yang2026explainable}: An explainable token-filtering baseline that scores tokens by task relevance, novelty, or reasoning-related criteria.

    \item \textbf{ssTOKEN}~\citep{qin2026sstoken}: A selective response-token training baseline under a fixed retained-token budget.

    \item \textbf{Wise-FT}~\citep{Wortsman_2022_CVPR}: A forgetting-resistant baseline that interpolates pre-trained and fine-tuned weights.

    \item \textbf{FLOW}~\citep{sanyal2025upweighting}: A conservative fine-tuning baseline that upweights easier or more stable samples to reduce forgetting.

    \item \textbf{TALR}~\citep{lin2026sft}: A token-level reweighting baseline that suppresses unstable token updates during fine-tuning.
\end{itemize}

\paragraph{Preference-alignment baselines.}
For preference optimization, all methods start from the same UltraChat-200k SFT warm-start checkpoint and train on the same UltraFeedback preference data.

\begin{itemize}[leftmargin=*]
    \item \textbf{Base}: The UltraChat-200k warm-started model before preference optimization.

    \item \textbf{DPO}~\citep{rafailov2023direct}: The standard sequence-level preference-optimization baseline.

    \item \textbf{ConfPO}~\citep{yoon2025confpo}: A confidence-aware preference-optimization baseline that downweights uncertain preference signals.

    \item \textbf{SePO}~\citep{yang2025selective}: A selective preference-optimization baseline that applies preference gradients to more informative response parts.

    \item \textbf{TI-DPO}~\citep{yang2026tokenimportance}: A token-importance-based DPO baseline that reweights or selects influential tokens in preference pairs.
\end{itemize}

AlphaToken differs from these baselines by estimating response-token value through gradient alignment, decomposing it into adaptation and stability terms, and further separating direct and causal-path effects.

\subsection{Training Data Details}
\label{app:exp:training_data}

We use separate training data for supervised fine-tuning and preference optimization. 
For supervised fine-tuning, we use Magicoder~\citep{wei2024magicoder} as the target-task corpus. 
Magicoder is an instruction-code dataset designed for code generation, where each instance contains a natural-language programming instruction and a reference solution. 
We train SFT models by applying the standard causal language-modeling objective only on response tokens, with AlphaToken or the corresponding baseline deciding which response-token losses are retained when token selection is used.

For preference optimization, we follow a two-stage training pipeline. 
First, each pre-trained backbone is warm-started with supervised fine-tuning on UltraChat-200k~\citep{ding2023enhancing}, which provides multi-turn instruction-following conversations and converts the base model into a chat-capable reference model. 
Second, preference optimization is performed on UltraFeedback~\citep{notus2023}. 
We use the binarized preference format, where each instance pairs a prompt with a preferred response and a rejected response, enabling DPO-style objectives to increase the relative likelihood of the preferred response over the rejected one. 
All preference-alignment baselines and AlphaToken use the same warm-start checkpoint and preference-training split. 
Table~\ref{tab:app_training_data} summarizes the training datasets used in our experiments.

\begin{table}[t]
\centering
\caption{Summary of training datasets used in our experiments.}
\label{tab:app_training_data}
\scriptsize
\setlength{\tabcolsep}{4.5pt}
\renewcommand{\arraystretch}{1.08}
\resizebox{\linewidth}{!}{
\begin{tabular}{l|l|l}
\toprule
\textbf{Dataset} & \textbf{Usage} & \textbf{Size} \\
\midrule
Magicoder
& SFT target-task training for code generation 
& 75K instructions \\
UltraChat-200k
& SFT warm-start for preference optimization 
& 200K conversations \\
UltraFeedback
& Preference optimization 
& 63.6K preference pairs \\
\bottomrule
\end{tabular}
}
\vspace{-0.05in}
\end{table}

\subsection{Evaluation Details}
\label{app:exp:evaluation}

We evaluate model performance from two perspectives: general capability retention and post-training target performance. 
For supervised fine-tuning, HumanEval~\citep{chen2021evaluating} is used as the target-task benchmark for code generation, while ARC-C~\citep{clark2018think}, HellaSwag~\citep{zellers2019hellaswag}, MMLU~\citep{hendrycks2021measuring}, and GSM8K~\citep{cobbe2021training} evaluate general capability retention. 
For preference optimization, AlpacaEval~2~\citep{alpaca_eval} and Arena-Hard~v0.1~\citep{arenahard2024} are used to evaluate instruction-following preference performance, while the same four general benchmarks, ARC-C~\citep{clark2018think}, HellaSwag~\citep{zellers2019hellaswag}, MMLU~\citep{hendrycks2021measuring}, and GSM8K~\citep{cobbe2021training}, are used to measure retention. 
All reported numbers are percentages. 
General Capability Avg. is the arithmetic mean over ARC-C, HellaSwag, MMLU, and GSM8K. 
For SFT, Overall is the arithmetic mean of General Capability Avg. and HumanEval PASS@1. 
For preference optimization, Preference Avg. is the arithmetic mean of AlpacaEval~2 and Arena-Hard~v0.1, and Overall is the arithmetic mean of General Capability Avg. and Preference Avg. 
Table~\ref{tab:app_eval_benchmarks} summarizes the evaluation benchmarks and metrics.

\begin{table}[t]
\centering
\caption{Evaluation benchmarks and detailed settings.}
\label{tab:app_eval_benchmarks}
\scriptsize
\setlength{\tabcolsep}{3.5pt}
\renewcommand{\arraystretch}{1.05}
\resizebox{\linewidth}{!}{
\begin{tabular}{l|l|l|l}
\toprule
\textbf{Benchmark} & \textbf{Aspect} & \textbf{Metric} & \textbf{Setting} \\
\midrule
ARC-C & General capability retention & Accuracy & 0-shot \\
HellaSwag & General capability retention & Accuracy & 0-shot \\
MMLU & General capability retention & Accuracy & 5-shot \\
GSM8K & General capability retention & Exact-match accuracy & 5-shot \\
HumanEval & SFT target-task adaptation & PASS@1 & 0-shot \\
AlpacaEval~2 & Preference optimization & Win rate & Standard judge \\
Arena-Hard~v0.1 & Preference optimization & Win rate & Standard judge \\
\bottomrule
\end{tabular}
}
\vspace{-0.05in}
\end{table}

\begin{itemize}[leftmargin=*]
     \item \textbf{ARC-C}~\citep{clark2018think}: ARC-Challenge is a multiple-choice science QA benchmark for grade-school scientific reasoning. Each question requires selecting the correct answer from several candidates. We report Accuracy.

    \item \textbf{HellaSwag}~\citep{zellers2019hellaswag}: HellaSwag evaluates commonsense natural language inference. Given an event description or partial context, the model selects the most plausible continuation from candidates. We report Accuracy.

    \item \textbf{MMLU}~\citep{hendrycks2021measuring}: Massive Multitask Language Understanding evaluates knowledge and reasoning across domains, including STEM, humanities, social sciences, and professional knowledge. We report 5-shot Accuracy.

    \item \textbf{GSM8K}~\citep{cobbe2021training}: GSM8K evaluates mathematical reasoning on grade-school word problems. The model must generate a numerical answer after multi-step reasoning. We report exact-match accuracy after answer extraction.
    
    \item \textbf{HumanEval}~\citep{chen2021evaluating}: HumanEval evaluates correctness for code generation. Each problem provides a Python function signature and natural-language description, and the solution is judged by hidden unit tests. We report PASS@1 with one solution per problem.

    \item \textbf{AlpacaEval~2}~\citep{alpaca_eval}: AlpacaEval~2 evaluates instruction-following through pairwise automatic judging. The judge compares model responses against reference or baseline responses under the official protocol. We report win rate.

    \item \textbf{Arena-Hard~v0.1}~\citep{arenahard2024}: Arena-Hard~v0.1 is a challenging instruction-following benchmark designed to better separate strong aligned models. It uses pairwise automatic judging on hard user queries. We report win rate.
\end{itemize}

We use the following metrics in our experiments.

\begin{itemize}[leftmargin=*]
    \item \textbf{Accuracy (ACC)} measures the proportion of correct predictions across examples. For multiple-choice benchmarks, the prediction is correct if the selected option matches the reference answer:
    \begin{equation}
    \mathrm{ACC}
    =
    \frac{1}{N}
    \sum_{i=1}^{N}
    \mathbbm{1}\!\left[\hat{y}_i \in \mathcal{Y}_i\right],
    \end{equation}
    where $N$ is the number of evaluation examples, $\hat{y}_i$ is the model prediction, and $\mathcal{Y}_i$ is the set of acceptable answers for the $i$-th example.

    \item \textbf{Exact Match (EM)} measures whether the normalized prediction exactly matches the normalized reference answer:
    \begin{equation}
    \mathrm{EM}
    =
    \frac{1}{N}
    \sum_{i=1}^{N}
    \mathbbm{1}\!\left[\mathrm{norm}(\hat{y}_i)=\mathrm{norm}(y_i)\right],
    \end{equation}
    where $y_i$ is the reference answer and $\mathrm{norm}(\cdot)$ denotes the benchmark-specific answer normalization function.

    \item \textbf{PASS@1} measures whether a single generated program passes all unit tests for a HumanEval problem:
    \begin{equation}
    \mathrm{PASS@1}
    =
    \frac{1}{N}
    \sum_{i=1}^{N}
    \mathbbm{1}\!\left[\mathrm{tests}(\hat{c}_i)=1\right],
    \end{equation}
    where $\hat{c}_i$ is the generated code for the $i$-th problem and $\mathrm{tests}(\hat{c}_i)=1$ indicates that all associated unit tests are passed.

    \item \textbf{Win Rate (WR)} measures the pairwise preference performance on AlpacaEval~2 and Arena-Hard~v0.1:
    \begin{equation}
    \mathrm{WR}
    =
    \frac{N_{\mathrm{win}}+0.5\,N_{\mathrm{tie}}}{N},
    \end{equation}
    where $N_{\mathrm{win}}$ and $N_{\mathrm{tie}}$ denote the number of wins and ties assigned by the automatic judge, and $N$ is the total number of judged examples.
\end{itemize}

\subsection{Implementation Details}
\label{app:exp:setup}

All methods are implemented under a unified post-training setup and optimized with AdamW using a cosine learning-rate scheduler. 
We use bfloat16 mixed-precision training and gradient checkpointing for all backbones. 
For supervised fine-tuning on Magicoder~\citep{wei2024magicoder}, we use a global batch size of 64, a maximum sequence length of 4096, and train for 3 epochs. 
For preference optimization, we first warm-start each pre-trained backbone on UltraChat-200k~\citep{ding2023enhancing} with a global batch size of 128, a maximum sequence length of 2048, and 2 epochs of SFT. 
We then perform preference optimization on UltraFeedback~\citep{notus2023} with a global batch size of 64 preference pairs, maximum prompt length 1024, maximum response length 1024, and 3 training epochs. 
The peak learning rate is $2{\times}10^{-5}$ for Llama-3.2-3B and
Gemma-3-4B, and $1{\times}10^{-5}$ for Qwen-3.5-9B.

For preference optimization training, the SFT warm-start checkpoint is used as the reference model.
We set the DPO temperature coefficient to $\beta=0.1$.
The peak learning rate is $2{\times}10^{-6}$ for Llama-3.2-3B and
Gemma-3-4B, and $1{\times}10^{-6}$ for Qwen-3.5-9B. Hyperparameters are selected according to the target validation split and used for the final run.
The same tokenizer, chat template, optimizer, scheduler, batch construction, and evaluation protocol are used for all methods under the same backbone. Package versions, preprocessing scripts, decoding parameters, judge settings, and answer-extraction rules are provided in the anonymous code repository.

For LoRA~\citep{hu2022lora}, we insert low-rank adapters into the attention and MLP projection layers and tune only the adapter parameters. 
For data-selection and token-selection baselines, including LESS~\citep{xia2024less}, Token Cleaning~\citep{pang2025token}, STM~\citep{wu2025mitigating}, XTF~\citep{yang2026explainable}, ssTOKEN~\citep{qin2026sstoken}, ConfPO~\citep{yoon2025confpo}, SePO~\citep{yang2025selective}, and TI-DPO~\citep{yang2026tokenimportance},  we follow the official implementation and recommended settings.  
When a method requires a retained-token ratio, we use the same retained ratio as AlphaToken to ensure a matched effective token budget. 
All experiments are conducted on $4{\times}$NVIDIA A100 GPUs.

\subsection{Training and Validation Setup}
\label{app:exp:alphatoken_setting}

AlphaToken requires two auxiliary signals during training: a target-side validation gradient and a data-free retention proxy. 
For the target-side signal, we hold out a validation subset of size
\(B_{\mathrm{val}}=32\) from the training corpus.
For SFT, this subset is sampled from Magicoder~\citep{wei2024magicoder}; for preference optimization, it is sampled from UltraFeedback~\citep{notus2023}. 
These validation examples are used only to compute the target-side gradient signal in Eq.~\eqref{eq:phi_tgt_dir} and Eq.~\eqref{eq:phi_tgt_causal}; they are excluded from parameter optimization.

During AlphaToken training, we concatenate the fixed target-validation mini-batch with each training mini-batch and obtain the validation-side activation and error signals in the same forward/backward pass. 
The validation-side token signals are averaged over validation tokens to keep
the scale independent of \(B_{\mathrm{val}}\).
By default, AlphaToken scores response tokens using the last $K=3$ Transformer layers, causal window $W=32$, retained-token ratio $\rho=0.5$, and stability weight $\lambda=1.5$. 
The threshold $\tau_{\rho}$ is computed within each training batch, so that each batch retains the top-$\rho$ response tokens according to the composite value in Eq.~\eqref{eq:phi_total}.

For the retention proxy, we construct the diagonal Monte-Carlo Fisher once before training from 1{,}000 prompts sampled from the current training distribution. 
For SFT, the reference checkpoint $\bm{\theta}_{\mathrm{ref}}$ is the original pre-trained backbone. 
For preference optimization, $\bm{\theta}_{\mathrm{ref}}$ is the UltraChat-200k warm-start checkpoint before DPO optimization. 
Labels used in the Monte-Carlo Fisher are self-sampled from $p_{\bm{\theta}_{\mathrm{ref}}}$, so no retention validation data or old-task examples are required. 
The resulting diagonal Fisher is cached throughout training. 
At each step, AlphaToken refreshes the layer-wise proxy slice
$\bm{V}^{l}=\bm{F}_{\bm{W}^{l}}\odot(\bm{W}^{l}-\bm{W}_{\mathrm{ref}}^{l})$
and computes the activation--parameter contraction in Eq.~\eqref{eq:ap_gdp}.

For Alpha-DPO, token values are computed separately for the chosen and rejected responses using the DPO-scaled token-level error signals. 
The sequence-level DPO coefficient is computed from the unmasked preference logit and detached from the token mask. 
The chosen and rejected branches use separate within-branch top-$\rho$ thresholds, producing masks $m_t^+$ and $m_t^-$. 
These masks are detached before the final loss computation, so AlphaToken changes only which response-token losses receive gradients and does not introduce second-order gradients through the valuation procedure.

\section{Extended Experimental Analysis}

\subsection{Analysis of the Fisher-Drift Proxy}
\label{app:fisher_proxy_analysis}

AlphaToken replaces the inaccessible retention-validation gradient with a Fisher-drift proxy. 
Rather than exactly reproducing the oracle retention gradient, the proxy is expected to rank response tokens whose path-aware gradients are compatible with preserving the reference model. 
For diagnosis, we build a small oracle retention set with 32 held-out examples, including 8 examples from each of ARC-C, HellaSwag, MMLU, and GSM8K. 
These examples are used only for this analysis and are excluded from training, Fisher construction, hyperparameter tuning, and model selection.

For each response token, we compare its Fisher proxy score with its oracle retention-gradient score computed from the held-out retention set. 
Since raw gradient inner products are scale-dependent and heavy-tailed, both scores are linearly normalized to \([-1,1]\) for visualization only, preserving the token ranking used in the Top-\(\rho\) analysis.

\begin{figure}[t]
    \centering
    \includegraphics[width=0.98\linewidth]{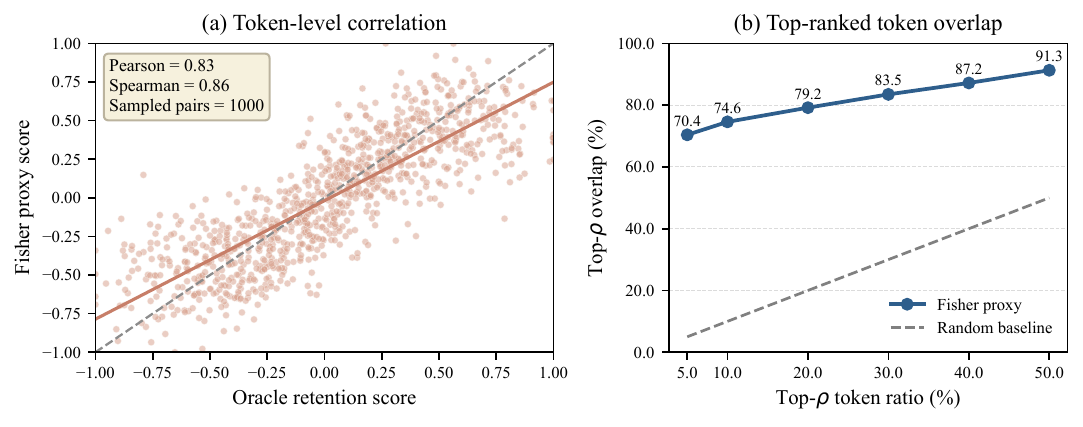}
    \caption{Left: correlation between normalized proxy and oracle retention scores. 
Right: Top-\(\rho\) overlap between proxy- and oracle-selected tokens.
    }
    \label{fig:fisher_proxy_fidelity}
\end{figure}

As shown in Fig.~\ref{fig:fisher_proxy_fidelity}, the Fisher proxy is strongly correlated with oracle retention-gradient scores and achieves much higher Top-\(\rho\) overlap than random selection, supporting its use as a reliable ranking signal for retention-aware token selection.

We further study whether the diagonal Fisher approximation used in AlphaToken is sufficient for the data-free retention proxy on Llama-3.2-3B/SFT. 
Table~\ref{tab:app:fisher_structure} compares the default diagonal Fisher with a more expensive Kronecker-factored approximation, which preserves layerwise input-output correlations in the scoring layers. 
All other settings, including the retained-token ratio, causal window, scoring layers, validation batch, and training protocol, are kept fixed.

\begin{table}[t]
\centering
\scriptsize
\caption{Fisher comparison on Llama-3.2-3B/SFT.}
\label{tab:app:fisher_structure}
\setlength{\tabcolsep}{5pt}
\renewcommand{\arraystretch}{0.95}
\resizebox{\linewidth}{!}{
\begin{tabular}{@{}l|ccc|c@{}}
\toprule
\textbf{Variant} & \textbf{Retention Avg.} & \textbf{HE} & \textbf{Overall} & \textbf{Rel. Time} \\
\midrule
No Fisher proxy & 43.18 & \textbf{44.36} & 43.77 & \textbf{0.94$\times$} \\
Diagonal Fisher & 45.47 & 43.98 & 44.73 & 1.00$\times$ \\
Kronecker Fisher & \textbf{45.56} & 44.02 & \textbf{44.79} & 1.28$\times$ \\
\bottomrule
\end{tabular}}
\end{table}

The Kronecker Fisher brings only a small improvement over the diagonal approximation while substantially increasing scoring cost. 
In contrast, diagonal Fisher achieves nearly the same retention and overall performance with much lower overhead. 
Removing the Fisher proxy yields slightly higher target adaptation but noticeably weaker retention, confirming the importance of the Fisher-weighted drift signal. 
These results support diagonal Fisher as a practical default for AlphaToken.
\subsection{Value-Propagation Approximation Diagnostics}
\label{app:exp:value_path_fidelity}

\begin{figure}[t]
\centering
\includegraphics[width=\linewidth]{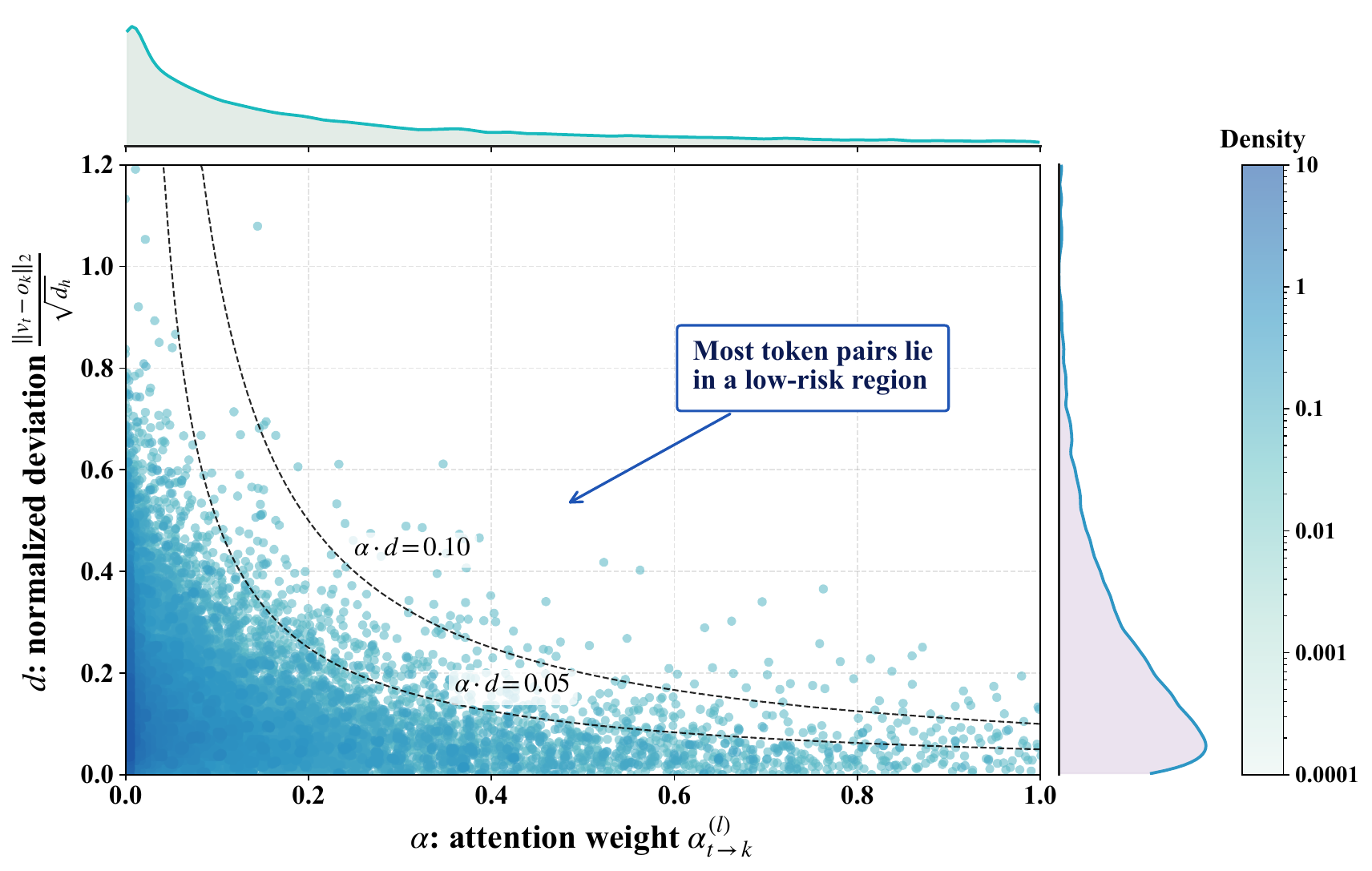}
\caption{Value-Propagation approximation diagnostics. We visualize the joint distribution of attention weight $\alpha_{t\to k}^{l}$ and normalized value-output deviation $d=\|\bm{v}_t-\bm{o}_k\|_2/\sqrt{d_h}$ over causal-window token pairs. Marginal density curves are shown on the top and right. Dashed curves indicate constant approximation-risk levels measured by $\alpha\cdot d$. Most token pairs concentrate in low-risk regions, suggesting that the omitted score-propagation term is typically limited.}
\vspace{-0.1in}
\label{fig:app:vp_fidelity}
\end{figure}

Figure~\ref{fig:app:vp_fidelity} provides an empirical diagnostic for the Value-Propagation approximation used in the causal path. We collect causal-window token pairs from the scoring layers during training and visualize the joint distribution of the attention weight $\alpha_{t\to k}^{l}$ and the normalized value-output deviation $d=\|\bm{v}_t-\bm{o}_k\|_2/\sqrt{d_h}$. The top and right marginal curves show that both quantities are highly skewed toward small values. More importantly, most token pairs concentrate in the lower-left region of the joint space, indicating that they either receive small attention weights, have small value-output deviation, or both.

The dashed curves mark levels of the product $\alpha\cdot d$, which serves as a proxy for the approximation-risk factor appearing in the Value-Propagation error bound. The distribution shows that high-risk pairs with large attention weight and large value-output deviation are rare, while most token pairs fall below the low-risk contours. This pattern supports Value-Propagation as a practical causal-path estimator: for most token pairs encountered during training, the residual score-dependent component remains concentrated in low-risk regions.

\subsection{Token-Coverage Effect of AlphaToken}
\label{app:exp:token_tsne}

We further examine whether the causal-path term changes the types of response tokens selected by AlphaToken. We visualize contextual embeddings under four policies: random selection, low-perplexity selection, direct-only valuation, and full AlphaToken, all retaining 50\% of tokens. For comparability, we fit one t-SNE projection on the union of selected embeddings and use the same two-dimensional coordinates for all policies.

\begin{figure}[t]
\centering
\includegraphics[width=\linewidth]{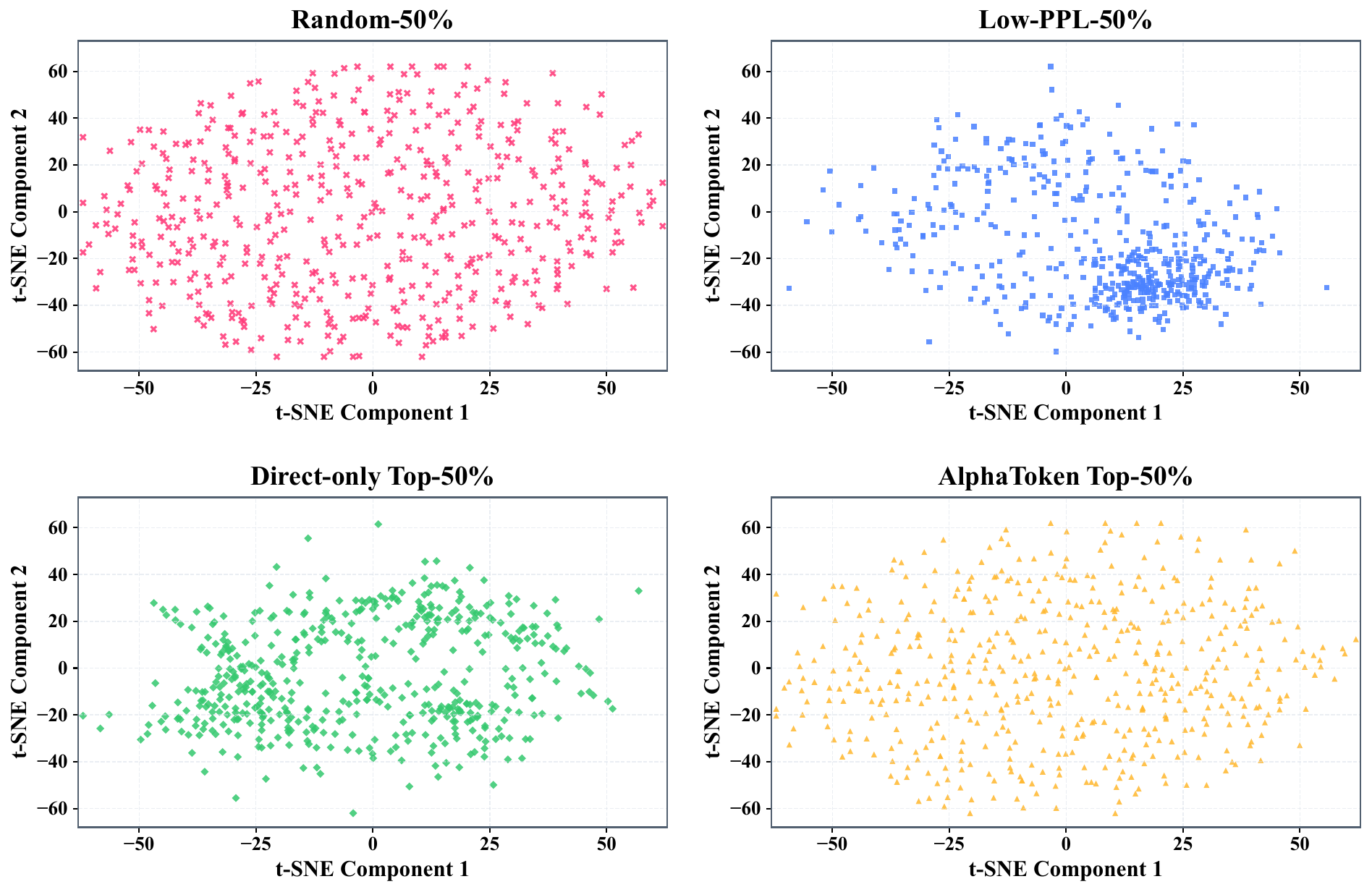}
\caption{t-SNE visualization of contextual embeddings of selected response tokens. 
All policies retain \(50\%\) of response tokens and use the same t-SNE space fitted on all selected token embeddings. 
Low-PPL selection concentrates in a narrower region, while Direct-only favors locally salient tokens. 
AlphaToken covers the space more evenly, suggesting that the causal-path term selects tokens with more diverse contextual roles.}
\label{fig:app:token_tsne}
\end{figure}

Figure~\ref{fig:app:token_tsne} shows that Low-PPL selection concentrates in a narrow embedding region, suggesting a preference for easy or fluent tokens. Direct-only valuation covers a broader region but remains visibly clustered, indicating that local signals may miss broader downstream-context roles. In contrast, AlphaToken distributes selected tokens more evenly, suggesting that the causal-path term recovers tokens with diverse contextual roles rather than only strong immediate signals. Random selection also covers a broad region, but unlike AlphaToken, it is not value-aware and does not distinguish tokens by their contribution to adaptation or retention.

\subsection{Retention-Probe Representation Stability}
\label{app:exp:representation_stability}
We examine whether AlphaToken better preserves the representation geometry of general-capability inputs after fine-tuning. We sample 1,000 HellaSwag prompts as retention probes and feed them into the pre-trained and fine-tuned models. For each prompt, we mean-pool the last-layer hidden states over non-padding tokens. For each backbone, we fit a shared t-SNE projection over pre-trained and post-fine-tuning representations, and visualize Standard FT and AlphaToken under the space.

\begin{figure*}[t]
\centering
\includegraphics[width=\linewidth]{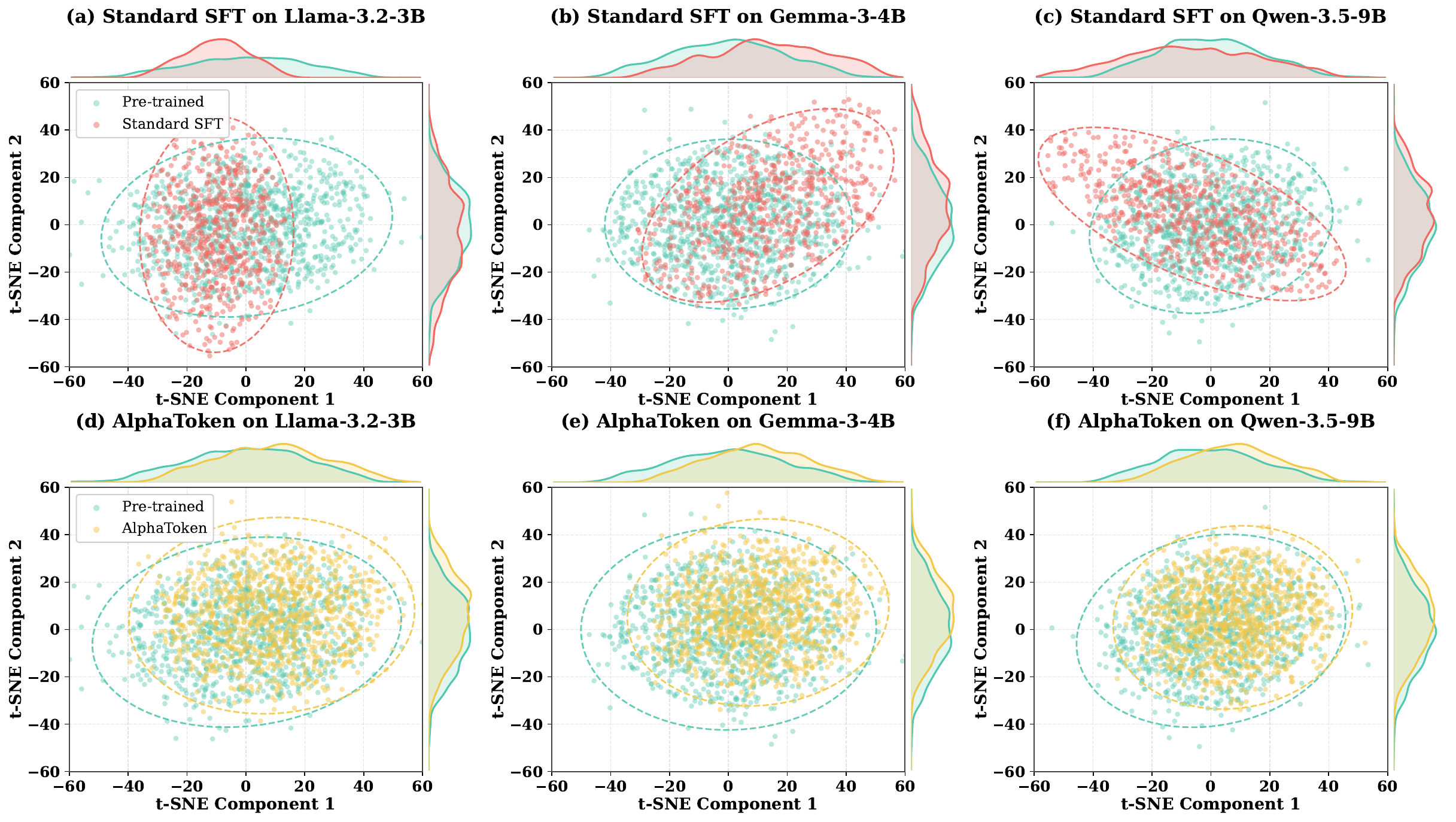}
\caption{Retention-probe representation stability. We visualize t-SNE distributions of pre-trained and post-fine-tuning prompt representations on 1,000 HellaSwag retention probes. Standard FT induces larger shifts, while AlphaToken keeps post-fine-tuning representations closer to the pre-trained distribution across three backbones.}
\label{fig:app:representation_stability}
\end{figure*}

Figure~\ref{fig:app:representation_stability} shows that Standard FT causes a clear distributional shift across all three backbones, reflected by displaced point clouds, marginal densities, and confidence ellipses. This suggests that direct target optimization substantially perturbs the representation geometry of retention probes. In contrast, AlphaToken produces post-fine-tuning representations that overlap more closely with the pre-trained distributions across Llama-3.2-3B, Gemma-3-4B, and Qwen-3.5-9B.

These results complement the main retention metrics. By selectively allocating gradient flow to high-value response tokens, AlphaToken reduces updates from less useful positions and better constrains representation-level drift, supporting its improved adaptation--retention trade-off.

\subsection{Qualitative Token-Level Visualizations}
\label{app:exp:additional_token_heatmaps}

To complement Sec.~\ref{subsec:component_attribution}, we provide additional qualitative visualizations on three SFT training examples. While Sec.~\ref{subsec:component_attribution} decomposes AlphaToken into four components on one dynamic-programming example, this appendix visualizes the final token value \(\Phi\) on more response fragments. Scores are normalized within each example, with darker colors indicating higher relative value.

\begin{figure*}[t]
\centering
\includegraphics[width=\linewidth]{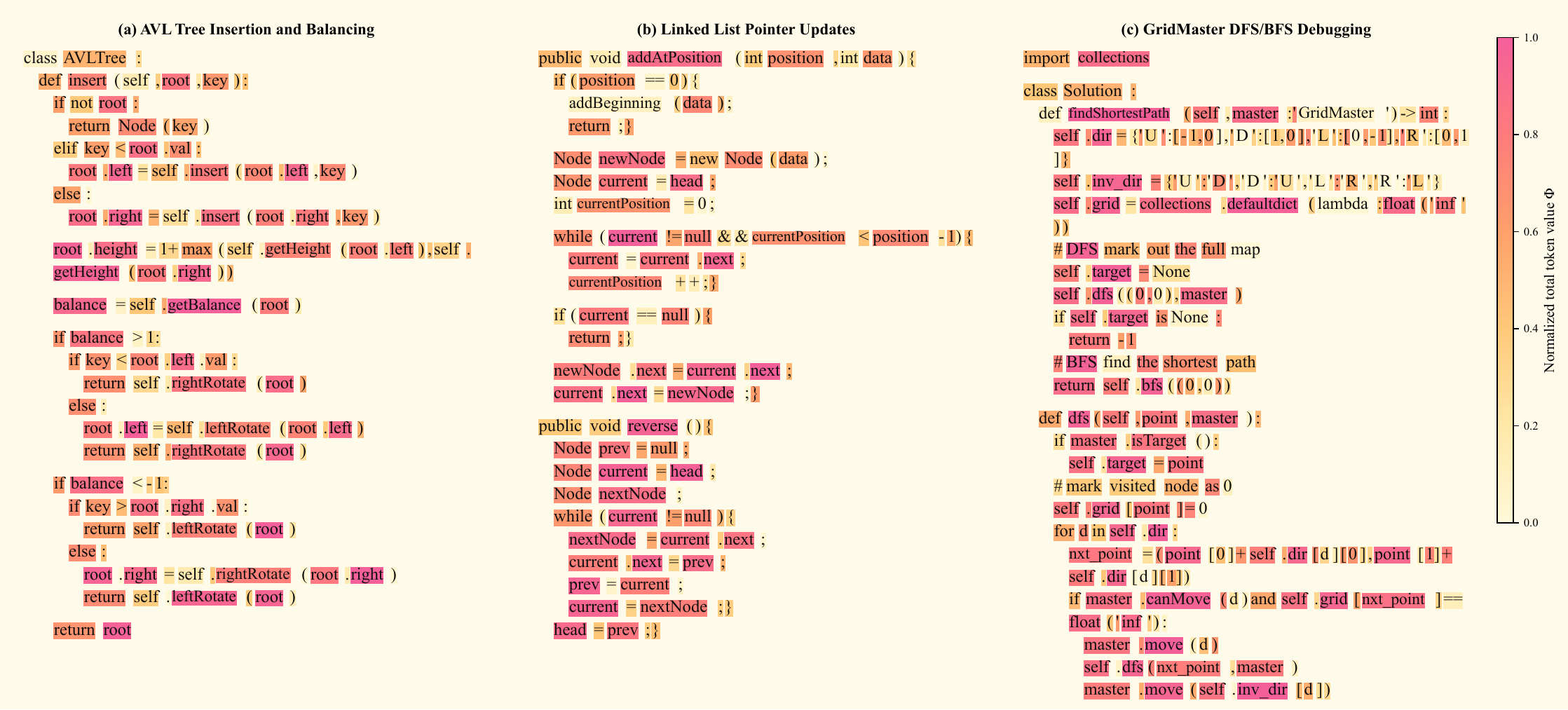}
\caption{Additional qualitative token-level visualizations on SFT examples. We visualize the final AlphaToken token value \(\Phi\) on three response fragments from the training data: (a) AVL-tree insertion and balancing, (b) linked-list pointer updates, and (c) GridMaster DFS/BFS debugging. Scores are normalized within each example. Darker colors indicate higher relative token value. AlphaToken assigns higher values to structurally important tokens, including balance checks and rotations, pointer updates, and DFS/BFS control-flow operations.}
\label{fig:app:sft_token_heatmaps}
\end{figure*}

Figure~\ref{fig:app:sft_token_heatmaps} shows that high-value tokens often align with structurally important program elements. In the AVL-tree example, they concentrate around height updates, balance-factor checks, and rotations such as \texttt{leftRotate} and \texttt{rightRotate}. In the linked-list example, AlphaToken highlights pointer-manipulation tokens for insertion and reversal, including \texttt{newNode.next}, \texttt{current.next}, \texttt{prev}, \texttt{nextNode}, and the final head update. In the GridMaster example, high-value tokens appear around DFS/BFS control flow, including direction maps, inverse moves, target detection, recursive exploration, and backtracking.

These examples suggest that \(\Phi\) is not uniformly distributed over tokens, but emphasizes tokens governing control flow, state transitions, pointer updates, and search behavior. This provides qualitative evidence that AlphaToken selects semantically and causally meaningful response positions across SFT coding tasks, rather than surface-level or uniformly distributed tokens.

\subsection{Running Time Analysis}
\label{app:exp:runtime}

We evaluate the efficiency of response-token selection methods on Llama-3.2-3B under matched hardware and training settings.
We focus on methods with explicit token scoring, selection, masking, or reweighting modules.
Train Time denotes the wall-clock time for one training epoch, including scoring or selection, mask construction, and parameter updates.
Peak Sel. Mem. records the maximum GPU memory during response-token scoring and mask construction.
SFT is measured on Magicoder, while preference optimization is measured on UltraFeedback after the shared UltraChat-200k warm-up.
One-time preprocessing, such as constructing and caching the Monte-Carlo Fisher, is excluded.

\begin{table}[t]
\centering
\small
\caption{
Efficiency profile of token selection methods in SFT on Magicoder with Llama-3.2-3B.
Train Time denotes one complete training epoch, and Peak Sel. Mem. denotes peak GPU memory during response-token scoring and mask construction.
}
\vspace{-0.1in}
\begin{tabular}{lcc}
\toprule
\textbf{Method} & \textbf{Train Time} & \textbf{Peak Sel. Mem.} \\
 & \textbf{(h/epoch)} & \textbf{(GB)} \\
\midrule
Token Cleaning  & 5.92 & 18.74 \\
STM             & 6.31 & 21.58 \\
XTF             & 6.57 & 24.36 \\
ssTOKEN         & 6.48 & 23.91 \\
AlphaToken      & 7.34 & 28.62 \\
\bottomrule
\end{tabular}
\label{tab:runtime_sft}
\vspace{-0.1in}
\end{table}

Table~\ref{tab:runtime_sft} reports the SFT efficiency profile.
Token Cleaning is the cheapest because it mainly uses local discrepancy signals.
STM, XTF, and ssTOKEN add token-level scoring or filtering, leading to moderately higher time and memory.
AlphaToken further introduces path-aware adaptation and stability valuation, with memory mainly from target-side validation signals and activation--parameter contraction for the Fisher-drift proxy.
Its runtime remains practical while providing stronger adaptation--retention performance.

\begin{table}[t]
\centering
\small
\caption{
Efficiency profile of response-token selection methods in preference optimization on UltraFeedback with Llama-3.2-3B.
Train Time denotes one complete training epoch, and Peak Sel. Mem. denotes peak GPU memory during scoring and mask construction over chosen and rejected responses.
}
\vspace{-0.1in}
\begin{tabular}{lcc}
\toprule
\textbf{Method} & \textbf{Train Time} & \textbf{Peak Sel. Mem.} \\
 & \textbf{(h/epoch)} & \textbf{(GB)} \\
\midrule
ConfPO      & 5.58 & 18.96 \\
SePO        & 6.24 & 22.47 \\
TI-DPO      & 6.43 & 23.68 \\
AlphaToken  & 7.72 & 30.14 \\
\bottomrule
\end{tabular}
\label{tab:runtime_po}
\end{table}

Table~\ref{tab:runtime_po} reports the preference-optimization efficiency profile.
ConfPO has the lowest cost because its selection signal relies on policy confidence.
SePO and TI-DPO add token-level preference filtering or importance estimation.
AlphaToken scores both chosen and rejected responses with target-alignment and stability-proxy signals, introducing extra valuation cost.
The overhead remains moderate relative to preference-token selection baselines, and the scoring memory fits within A100-class GPUs.
Overall, AlphaToken achieves a practical runtime and memory profile while improving the adaptation--retention trade-off.

\subsection{More Datasets and Baselines}
\label{app:more_datasets_baselines}

We further evaluate AlphaToken under an additional target domain and stronger forgetting-resistant baselines.
Table~\ref{tab:metamathqa-gemma4b} reports results on MetaMathQA, where GSM8K is the target benchmark and HumanEval is part of general-capability retention.
Standard FT achieves the highest GSM8K score but suffers a clear drop in general capability.
In contrast, AlphaToken obtains the best Overall score and General Capability Avg., while maintaining the second-best GSM8K performance.
This shows that AlphaToken better balances mathematical adaptation and retained general capabilities.

Table~\ref{tab:magicoder_forgetting_gemma4b} compares AlphaToken with forgetting-resistant baselines on Magicoder. 
Wise-FT and FLOW improve retention over Standard FT, and TALR gives the strongest competing trade-off. 
AlphaToken still achieves the best Overall score, outperforming TALR by $0.95$ points, and also obtains the highest HumanEval and General Capability Avg. 
These results indicate that path-aware response-token valuation remains effective against dedicated forgetting-mitigation methods.

\begin{table*}[t] 
\centering 
\caption{MetaMathQA fine-tuning results on Gemma-3-4B. The target task is GSM8K. HumanEval (HE) is reported as part of general capability to assess code-generation retention. Pre-trained models are excluded from color ranking. The best and second-best results are highlighted in \textcolor{red}{red} and \textcolor{blue}{blue}, respectively.} 
\vspace{-0.1in} 
\label{tab:metamathqa-gemma4b} 
\scriptsize 
\renewcommand{\arraystretch}{1.0} 
\setlength{\tabcolsep}{6pt} 
\definecolor{highlightpurple}{HTML}{E6E6FA} 
\begin{tabular}{c|c|ccccc|c|c} 
\toprule 
\multirow{2}{*}[-0.6ex]{\textbf{Model}} & \multirow{2}{*}[-0.6ex]{\textbf{Method}} & \multicolumn{5}{c|}{\textbf{General Capability Acc.~(\%)}} & \textbf{Target~(\%)} & \textbf{Overall} \\ 
\cmidrule(lr){3-7}\cmidrule(lr){8-8}\cmidrule(lr){9-9} 
& & \textbf{ARC-C} & \textbf{HellaSwag} & \textbf{MMLU} & \textbf{HE} & \textbf{Avg.} & \textbf{GSM8K} & \textbf{Avg.} \\ 
\midrule 
& Pre-trained & 51.47$_{\pm 0.00}$ & 57.36$_{\pm 0.00}$ & 59.58$_{\pm 0.00}$ & 35.40$_{\pm 0.00}$ & 50.95$_{\pm 0.00}$ & 36.96$_{\pm 0.00}$ & 43.96$_{\pm 0.00}$ \\ 
& Standard FT & 43.52$_{\pm 0.45}$ & 51.40$_{\pm 0.38}$ & 52.96$_{\pm 0.40}$ & 28.90$_{\pm 0.49}$ & 44.20$_{\pm 0.22}$ & \textcolor{red}{67.65$_{\pm 0.41}$} & 55.93$_{\pm 0.24}$ \\ 
& LoRA \venuetag{ICLR 2022} & 48.96$_{\pm 0.36}$ & 54.88$_{\pm 0.28}$ & 55.92$_{\pm 0.32}$ & 29.93$_{\pm 0.51}$ & 47.42$_{\pm 0.19}$ & 62.45$_{\pm 0.37}$ & 54.94$_{\pm 0.21}$ \\ 
& LESS \venuetag{ICML 2024} & 49.97$_{\pm 0.34}$ & \textcolor{blue}{55.94$_{\pm 0.30}$} & 55.34$_{\pm 0.29}$ & 30.42$_{\pm 0.44}$ & 47.92$_{\pm 0.17}$ & 65.42$_{\pm 0.38}$ & \textcolor{blue}{56.67$_{\pm 0.21}$} \\ 
& Token Cleaning \venuetag{ICML 2025} & \textcolor{blue}{50.15$_{\pm 0.38}$} & 55.31$_{\pm 0.35}$ & 55.72$_{\pm 0.34}$ & 30.58$_{\pm 0.44}$ & 47.94$_{\pm 0.19}$ & 63.20$_{\pm 0.40}$ & 55.57$_{\pm 0.22}$ \\ 
& STM \venuetag{NeurIPS 2025} & 49.53$_{\pm 0.40}$ & 55.70$_{\pm 0.34}$ & 56.40$_{\pm 0.27}$ & \textcolor{blue}{31.66$_{\pm 0.51}$} & \textcolor{blue}{48.32$_{\pm 0.19}$} & 64.15$_{\pm 0.39}$ & 56.24$_{\pm 0.22}$ \\ 
& XTF \venuetag{ICLR 2026} & 48.56$_{\pm 0.39}$ & 54.05$_{\pm 0.35}$ & 55.80$_{\pm 0.33}$ & 30.80$_{\pm 0.52}$ & 47.30$_{\pm 0.20}$ & 61.92$_{\pm 0.37}$ & 54.61$_{\pm 0.22}$ \\ 
& ssTOKEN \venuetag{ICLR 2026} & 49.74$_{\pm 0.37}$ & 54.66$_{\pm 0.31}$ & \textcolor{blue}{56.62$_{\pm 0.34}$} & 31.16$_{\pm 0.51}$ & 48.05$_{\pm 0.19}$ & 63.60$_{\pm 0.40}$ & 55.82$_{\pm 0.22}$ \\ 
\rowcolor{highlightpurple}\cellcolor{white} \multirow{-9}{*}{\rotatebox{90}{\textbf{Gemma-3-4B}}} & \textbf{AlphaToken} & \textcolor{red}{50.74$_{\pm 0.30}$} & \textcolor{red}{56.21$_{\pm 0.29}$} & \textcolor{red}{57.83$_{\pm 0.28}$} & \textcolor{red}{33.27$_{\pm 0.34}$} & \textcolor{red}{49.51$_{\pm 0.15}$} & \textcolor{blue}{66.18$_{\pm 0.39}$} & \textcolor{red}{57.85$_{\pm 0.19}$} \\ 
\bottomrule 
\end{tabular}  
\end{table*}

\begin{table*}[t]
\centering
\caption{
Magicoder fine-tuning results on Gemma-3-4B with forgetting-resistant baselines. 
Pre-trained models are excluded from color ranking. 
The best and second-best results are highlighted in \textcolor{red}{red} and \textcolor{blue}{blue}, respectively.
}
\vspace{-0.1in}
\label{tab:magicoder_forgetting_gemma4b}
\scriptsize
\renewcommand{\arraystretch}{1.0}
\setlength{\tabcolsep}{6pt}
\definecolor{highlightpurple}{HTML}{E6E6FA}
\begin{tabular}{c|c|ccccc|c|c}
\toprule
\multirow{2}{*}[-0.6ex]{\textbf{Model}} &
\multirow{2}{*}[-0.6ex]{\textbf{Method}} &
\multicolumn{5}{c|}{\textbf{General Capability Acc.~(\%)}} &
\textbf{Target~(\%)} &
\textbf{Overall} \\
\cmidrule(lr){3-7}\cmidrule(lr){8-8}\cmidrule(lr){9-9}
& & \textbf{ARC-C} & \textbf{HellaSwag} & \textbf{MMLU} & \textbf{GSM8K} & \textbf{Avg.} & \textbf{HE} & \textbf{Avg.} \\
\midrule
& Pre-trained
  & 51.45$_{\pm 0.00}$ & 56.86$_{\pm 0.00}$ & 59.60$_{\pm 0.00}$ & 37.00$_{\pm 0.00}$ & 51.23$_{\pm 0.00}$ & 35.36$_{\pm 0.00}$ & 43.30$_{\pm 0.00}$ \\
& Standard FT
  & 50.17$_{\pm 0.45}$ & 52.63$_{\pm 0.38}$ & 53.87$_{\pm 0.40}$ & 24.12$_{\pm 0.28}$ & 45.20$_{\pm 0.19}$ & 58.79$_{\pm 0.49}$ & 52.00$_{\pm 0.26}$ \\
& Wise-FT \venuetag{CVPR 2022}
  & 50.32$_{\pm 0.36}$ & 54.46$_{\pm 0.32}$ & 56.03$_{\pm 0.33}$ & 33.40$_{\pm 0.42}$ & 48.55$_{\pm 0.18}$ & 55.92$_{\pm 0.44}$ & 52.24$_{\pm 0.24}$ \\
& FLOW \venuetag{ICML 2025}
  & 50.83$_{\pm 0.34}$ & \textcolor{blue}{55.72$_{\pm 0.30}$} & 56.46$_{\pm 0.29}$ & 34.28$_{\pm 0.37}$ & 49.32$_{\pm 0.16}$ & 58.34$_{\pm 0.44}$ & 53.83$_{\pm 0.23}$ \\
& TALR \venuetag{ICLR 2026}
  & \textcolor{blue}{51.02$_{\pm 0.30}$} & \textcolor{red}{56.06$_{\pm 0.34}$} & \textcolor{blue}{57.88$_{\pm 0.25}$} & \textcolor{blue}{35.61$_{\pm 0.41}$} & \textcolor{blue}{50.14$_{\pm 0.17}$} & \textcolor{blue}{60.37$_{\pm 0.51}$} & \textcolor{blue}{55.26$_{\pm 0.27}$} \\
\rowcolor{highlightpurple}\cellcolor{white}
\multirow{-6}{*}{\rotatebox{90}{\textbf{Gemma-3-4B}}}
& \textbf{AlphaToken}
  & \textcolor{red}{51.10$_{\pm 0.30}$} & 55.12$_{\pm 0.29}$ & \textcolor{red}{58.22$_{\pm 0.28}$} & \textcolor{red}{36.58$_{\pm 0.46}$} & \textcolor{red}{50.26$_{\pm 0.17}$} & \textcolor{red}{62.15$_{\pm 0.34}$} & \textcolor{red}{56.21$_{\pm 0.19}$} \\
\bottomrule
\end{tabular}
\vspace{-0.20in}
\end{table*}

\subsection{Preference-Optimization Ablations}
\label{app:exp:pref_ablation}
Table~\ref{tab:app:pref_ablation_dpo} ablates AlphaToken under the preference-optimization setting on Gemma-3-4B. 
We evaluate three groups of design choices. 
First, we remove either the adaptation or stability objective to test whether the two valuation objectives are complementary. 
Second, we keep only the direct or causal path to test whether immediate token-level signals and downstream autoregressive influence provide non-redundant information. 
Third, we ablate two DPO-specific masking designs: using a shared threshold for chosen and rejected responses, and computing the DPO coefficient from the masked preference logit instead of the unmasked sequence-level logit.
\begin{table}[t]
\centering
\scriptsize
\caption{PO ablations on Gemma-3-4B.}
\vspace{-0.1in}
\label{tab:app:pref_ablation_dpo}
\setlength{\tabcolsep}{2.8pt}
\renewcommand{\arraystretch}{1.05}
\resizebox{\linewidth}{!}{
\begin{tabular}{@{}l|l|c|ccc|c@{}}
\toprule
\textbf{Ablation Axis} & \textbf{Variant} 
& \textbf{Gen.} 
& \textbf{AE2} 
& \textbf{A-Hard} 
& \textbf{Pref. Avg.} 
& \textbf{Overall} \\
\midrule
Full design 
& Full AlphaToken 
& 44.42 & 31.86 & 27.40 & 29.63 & \textbf{37.03} \\
\midrule
\multirow{2}{*}{Objective}
& Adaptation-only 
& 42.73 & \textbf{33.37} & \textbf{28.74} & \textbf{31.06} & 36.90 \\
& Stability-only 
& \textbf{45.08} & 25.42 & 21.37 & 23.40 & 34.24 \\
\midrule
\multirow{2}{*}{Path}
& Direct-only 
& 44.19 & 30.17 & 25.88 & 28.03 & 36.11 \\
& Causal-only 
& 43.81 & 28.46 & 24.59 & 26.53 & 35.17 \\
\midrule
\multirow{2}{*}{DPO masking}
& Shared $\tau_{\rho}$ for $y^{+}/y^{-}$ 
& 44.05 & 30.49 & 26.08 & 28.29 & 36.17 \\
& Masked DPO logit 
& 43.66 & 29.73 & 25.92 & 27.83 & 35.75 \\
\bottomrule
\end{tabular}}
\vspace{-0.05in}
\end{table}
The ablation results show three trends. 
First, the objective-axis variants expose the target--retention trade-off in preference optimization. 
Adaptation-only achieves the highest preference score, improving Preference Avg. from $29.63$ to $31.06$, but reduces General Capability Avg. from $44.42$ to $42.73$, indicating a stronger alignment tax when the retention proxy is removed. 
In contrast, Stability-only obtains the best retention score but substantially weakens preference learning, reducing Preference Avg. to $23.40$. 
Second, the path-axis variants show that direct and causal signals are complementary. 
Direct-only preserves more of the full model's performance than Causal-only, but both underperform the full design, suggesting that immediate token gradients and downstream causal influence capture different aspects of preference-relevant token value. 
Third, the DPO-specific ablations confirm that branch-aware masking is important. 
Using a shared threshold for chosen and rejected responses weakens preference learning, while computing the DPO coefficient from the masked logit makes the update less stable. 
These results support the Alpha-DPO design, where chosen and rejected responses are scored with separate within-branch thresholds and the sequence-level DPO coefficient is computed from the unmasked preference logit and detached before token-level masking.

\subsection{Preference Optimization Sensitivity}
\label{app:exp:dpo_sensitivity}

As shown in Table~\ref{tab:app:dpo_sensitivity}, AlphaToken remains robust under preference optimization across moderate hyperparameter ranges.
For the retained token ratio $\rho$, $\rho=0.4$ best preserves general capability, while $\rho=0.3$ underfits AE2 and Arena-Hard.
The default $\rho=0.5$ achieves the best Overall score.
Larger ratios reduce both preference and retention metrics, suggesting that low-value tokens introduce noisy gradients.

The stability weight $\lambda$ controls the balance between adaptation and retention.
Smaller values improve preference scores but reduce retention, whereas larger values preserve general capability at the cost of preference learning.
The best Overall score appears at $\lambda=1.5$, indicating that a moderately strong stability proxy reduces alignment tax without over-constraining optimization.

For the causal window $W$, performance improves from small windows and becomes stable after $W=32$.
We use $W=32$ as the default because it provides sufficient future token coverage while avoiding the extra cost of larger windows.
For scoring layers, larger last-$K$ generally improves performance, but the gain over last-$K=3$ is modest, making last-$K=3$ a practical choice for balancing efficiency and performance.
Similarly, increasing $B_{\mathrm{val}}$ improves valuation stability and becomes stable after $B_{\mathrm{val}}=32$.
We use $B_{\mathrm{val}}=32$ as the default to obtain stable target gradient estimates while avoiding the higher cost of larger validation batches.

The DPO coefficient $\beta$ controls preference update strength.
A small $\beta$ preserves general capability but under-optimizes preference learning, while larger values improve AE2 and Arena-Hard at the cost of retention.
The default $\beta=0.10$ ties for the best rounded Overall score while preserving better general capability than $\beta=0.20$.
Overall, AlphaToken is not overly sensitive to any hyperparameter, and the default setting balances preference learning, retention, and computational cost.
\begin{table*}[t]
\centering
\tiny
\caption{Parameter sensitivity of AlphaToken under preference optimization on Gemma-3-4B.}
\label{tab:app:dpo_sensitivity}
\setlength{\tabcolsep}{6.8pt}
\renewcommand{\arraystretch}{0.98}
\resizebox{\textwidth}{!}{
\begin{tabular}{@{}c|ccccc|ccc|c@{}}
\toprule
\textbf{Value}
& \multicolumn{5}{c|}{\textbf{General Capability Acc. (\%)}}
& \multicolumn{3}{c|}{\textbf{Preference Win Rate (\%)}}
& \multicolumn{1}{c}{\textbf{Overall}} \\
\cmidrule(lr){2-6}\cmidrule(lr){7-9}\cmidrule(lr){10-10}
& \textbf{ARC-C} & \textbf{HellaSwag} & \textbf{MMLU} & \textbf{GSM8K}
& \textbf{Avg.}
& \textbf{AE2} & \textbf{A-Hard}
& \textbf{Avg.}
& \textbf{Avg.} \\
\midrule
\rowcolor{gray!15}
\multicolumn{10}{c}{\textbf{Retained-token ratio $\rho$}} \\
0.3 & 50.61 & 51.53 & \textbf{52.45} & \textbf{24.45} & 44.76 & 20.31 & 16.90 & 18.61 & 31.68 \\
0.4 & \textbf{50.92} & \textbf{51.92} & 52.10 & 24.34 & \textbf{44.82} & 27.83 & 24.00 & 25.92 & 35.37 \\
0.5 & 50.48 & 51.46 & 51.68 & 24.06 & 44.42 & 31.86 & \textbf{27.40} & \textbf{29.63} & \textbf{37.03} \\
0.6 & 49.92 & 50.94 & 51.25 & 23.41 & 43.88 & \textbf{31.99} & 26.80 & 29.40 & 36.64 \\
0.7 & 49.57 & 50.72 & 51.41 & 22.92 & 43.66 & 29.94 & 24.40 & 27.17 & 35.41 \\
\midrule
\rowcolor{gray!15}
\multicolumn{10}{c}{\textbf{Stability weight $\lambda$}} \\
0.6 & 48.54 & 49.95 & 50.26 & 22.33 & 42.77 & \textbf{33.29} & \textbf{28.80} & \textbf{31.05} & 36.91 \\
0.9 & 49.61 & 50.68 & 50.95 & 23.80 & 43.76 & 32.11 & 27.90 & 30.01 & 36.88 \\
1.2 & 50.86 & 51.82 & 52.04 & 24.24 & 44.74 & 30.19 & 26.00 & 28.10 & 36.42 \\
1.5 & 50.48 & 51.46 & 51.68 & 24.06 & 44.42 & 31.86 & 27.40 & 29.63 & \textbf{37.03} \\
1.8 & \textbf{51.15} & \textbf{52.16} & \textbf{52.42} & \textbf{24.39} & \textbf{45.03} & 27.95 & 23.90 & 25.93 & 35.48 \\
\midrule
\rowcolor{gray!15}
\multicolumn{10}{c}{\textbf{Causal window $W$}} \\
4  & 50.05 & 51.04 & 51.20 & 23.79 & 44.02 & 29.25 & 25.00 & 27.13 & 35.57 \\
8  & 50.22 & 51.26 & 51.43 & \textbf{24.09} & 44.25 & 30.87 & 26.60 & 28.74 & 36.49 \\
16 & 50.46 & 51.43 & \textbf{51.70} & 24.01 & 44.40 & 31.93 & \textbf{27.50} & 29.72 & \textbf{37.06} \\
32 & \textbf{50.48} & \textbf{51.46} & 51.68 & 24.06 & \textbf{44.42} & 31.86 & 27.40 & 29.63 & 37.03 \\
64 & 50.42 & 51.40 & 51.69 & 23.97 & 44.37 & \textbf{31.99} & \textbf{27.50} & \textbf{29.75} & \textbf{37.06} \\
\midrule
\rowcolor{gray!15}
\multicolumn{10}{c}{\textbf{Last-$K$ scoring layers}} \\
1 & 49.98 & 50.92 & 51.15 & 23.79 & 43.96 & 30.12 & 25.70 & 27.91 & 35.94 \\
2 & 50.20 & 51.18 & 51.42 & 24.00 & 44.20 & 31.06 & 26.80 & 28.93 & 36.57 \\
3 & 50.48 & 51.46 & 51.68 & 24.06 & 44.42 & 31.86 & 27.40 & 29.63 & 37.03 \\
5 & 50.54 & 51.50 & 51.76 & \textbf{24.12} & 44.48 & 31.99 & 27.50 & 29.75 & 37.11 \\
8 & \textbf{50.58} & \textbf{51.56} & \textbf{51.82} & \textbf{24.12} & \textbf{44.52} & \textbf{32.05} & \textbf{27.60} & \textbf{29.83} & \textbf{37.17} \\
\midrule
\rowcolor{gray!15}
\multicolumn{10}{c}{\textbf{Target validation size $B_{\mathrm{val}}$}} \\
4  & 50.02 & 51.12 & 51.34 & 23.72 & 44.05 & 30.75 & 26.50 & 28.63 & 36.34 \\
8  & 50.27 & 51.30 & 51.54 & 23.97 & 44.27 & 31.24 & 27.00 & 29.12 & 36.70 \\
16 & 50.54 & 51.51 & 51.74 & 24.09 & 44.47 & \textbf{31.93} & \textbf{27.50} & \textbf{29.72} & 37.09 \\
32 & 50.48 & 51.46 & 51.68 & 24.06 & 44.42 & 31.86 & 27.40 & 29.63 & 37.03 \\
64 & \textbf{50.58} & \textbf{51.54} & \textbf{51.78} & \textbf{24.10} & \textbf{44.50} & \textbf{31.93} & \textbf{27.50} & \textbf{29.72} & \textbf{37.11} \\
\midrule
\rowcolor{gray!15}
\multicolumn{10}{c}{\textbf{DPO coefficient $\beta$}} \\
0.05 & \textbf{50.66} & \textbf{51.58} & \textbf{51.90} & \textbf{24.10} & \textbf{44.56} & 29.94 & 25.70 & 27.82 & 36.19 \\
0.10 & 50.48 & 51.46 & 51.68 & 24.06 & 44.42 & 31.86 & 27.40 & 29.63 & \textbf{37.03} \\
0.20 & 49.98 & 51.08 & 51.52 & 23.26 & 43.96 & 32.30 & 27.90 & 30.10 & \textbf{37.03} \\
0.30 & 49.32 & 50.64 & 50.96 & 22.68 & 43.40 & \textbf{32.55} & \textbf{28.00} & \textbf{30.28} & 36.84 \\
\bottomrule
\end{tabular}}
\vspace{-0.05in}
\end{table*}

\section{Artifact Licenses and Intended Use}
\label{app:artifact_use}

We use publicly available artifacts, including model checkpoints, training corpora, evaluation benchmarks, and baseline implementations, through their official release channels and in accordance with their licenses, terms of use, and intended research purposes. Magicoder, UltraChat-200K, and UltraFeedback are used for supervised fine-tuning, warm-start instruction tuning, and preference optimization, respectively, while HumanEval, AlpacaEval 2, Arena-Hard, ARC-C, HellaSwag, MMLU, and GSM8K are used only for evaluation. We do not redistribute original datasets, benchmarks, or model weights. Our released code contains implementation scripts, configurations, and reproduction instructions for research use only, and any derivative outputs are used solely for research analysis rather than deployment outside research contexts.

\end{document}